\documentclass[journal]{IEEEtran}
\ifCLASSINFOpdf
  \usepackage[pdftex]{graphicx}
\else
\fi
\usepackage{amssymb}
\usepackage[cmex10]{amsmath}
\usepackage{algorithm}
\usepackage{algorithmic}

\usepackage{cases}
\usepackage{array}

\usepackage{booktabs}

\usepackage[tight,footnotesize]{subfigure}
\begin{document}
%
\title{Multi-view Vector-valued Manifold Regularization for Multi-label Image Classification}
%
%
%

\author{Yong~Luo,
        Dacheng~Tao,~\IEEEmembership{Senior Member,~IEEE,}
        Chang~Xu,
        Chao~Xu,\\
        Hong~Liu,
        Yonggang~Wen,~\IEEEmembership{Member,~IEEE}
\thanks{Y. Luo, C. Xu, and C. Xu are with the Key Laboratory of Machine Perception (Ministry of Education),
School of Electronics Engineering and Computer Science, Peking University, Beijing, 100871, China.}
\thanks{D. Tao is with the Centre for Quantum Computation and Intelligent Systems,
University of Technology, Sydney, Jones Street, Ultimo, NSW 2007, Sydney, Australia.}
\thanks{H. Liu is with the Engineering Lab on Intelligent Perception for Internet of Things, Shenzhen Graduate School, Peking University, China. (Email: liuh@pkusz.edu.cn)}
\thanks{Y. Wen is with the Division of Networks and Distributed Systems, School of Computer Engineering, Nanyang Technological University, Singapore. (Email: ygwen@ntu.edu.sg)}
\thanks{\copyright 2013 IEEE. Personal use of this material is permitted. Permission from IEEE must be obtained for all other uses, in any current or future media, including reprinting/republishing this material for advertising or promotional purposes, creating new collective works, for resale or redistribution to servers or lists, or reuse of any copyrighted component of this work in other works.}
}

%
%


\markboth{$>$ \normalsize{TNNLS-2012-P-0852 R}\footnotesize{evision} \normalsize{2} $<$}%
{Shell \MakeLowercase{\textit{et al.}}: Bare Demo of IEEEtran.cls for Journals}

%



\maketitle

\begin{abstract}
In computer vision, image datasets used for classification are naturally associated with multiple labels and comprised of multiple views, because each image may contain several objects (e.g. pedestrian, bicycle and tree) and is properly characterized by multiple visual features (e.g. color, texture and shape). Currently available tools ignore either the label relationship or the view complementary. Motivated by the success of the vector-valued function that constructs matrix-valued kernels to explore the multi-label structure in the output space, we introduce multi-view vector-valued manifold regularization (MV$\mathbf{^3}$MR) to integrate multiple features. MV$\mathbf{^3}$MR exploits the complementary property of different features and discovers the intrinsic local geometry of the compact support shared by different features under the theme of manifold regularization. We conducted extensive experiments on two challenging, but popular datasets, PASCAL VOC' 07 (VOC) and MIR Flickr (MIR), and validated the effectiveness of the proposed MV$\mathbf{^3}$MR for image classification.
\end{abstract}

\begin{IEEEkeywords}
Image classification, semi-supervised, multi-label, multi-view, manifold
\end{IEEEkeywords}

%
\IEEEpeerreviewmaketitle

\section{Introduction}
\label{sec:Introduction}
\begin{figure*}[!t]
\centering
\includegraphics[width=2.0\columnwidth]{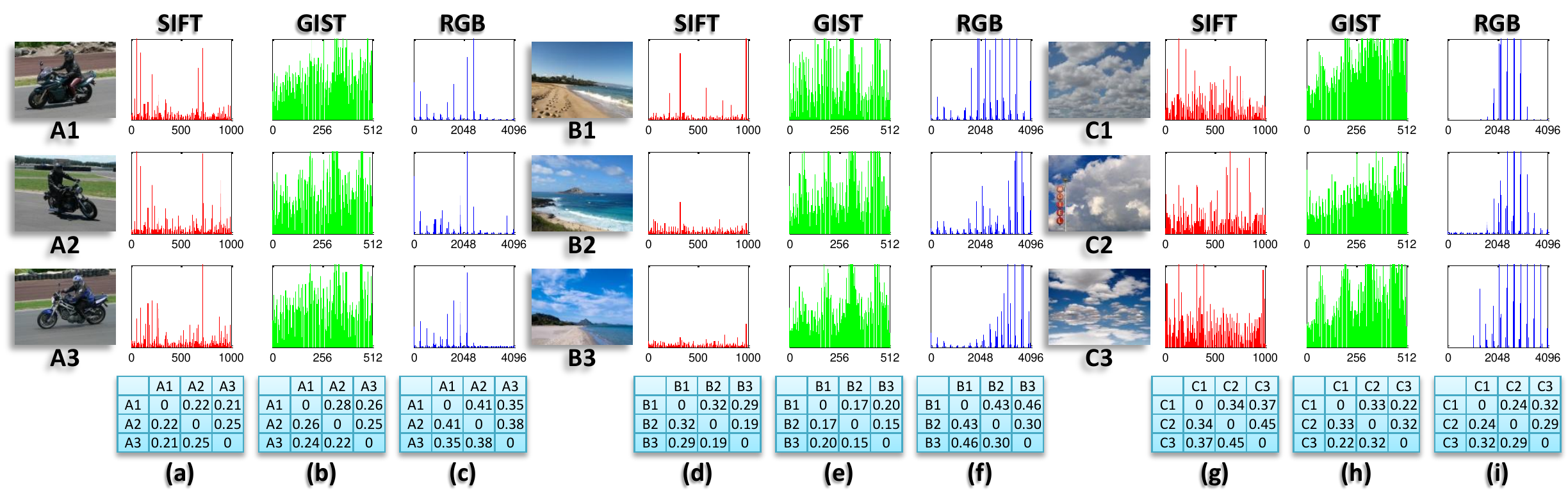}
\caption{A1-A3, B1-B3 and C1-C3 are images of person riding a motorbike, seaside and clouds in the sky respectively. Each (a)-(i) contains feature representations and a distance matrix of the samples in a particular view. All the distances have been normalized here.}
\label{fig:Example_Images}
\vspace{-5mm}
\end{figure*}

%
%
%
%
\IEEEPARstart{A} natural image can be summarized by several keywords (or labels). To conduct image classification by directly using binary classification methods \cite{Boutell-et-al-PR-2004, Guillaumin-et-al-CVPR-2010}, it is necessary to assume that labels are independent, although most labels appearing in one image are related to one another. Examples are given in Fig. \ref{fig:Example_Images}, where A1-A3 shows a ``person'' rides a ``motorbike'', B1-B3 indicates ``sea'' usually co-occurs with ``sky'' and C1-C3 shows some ``clouds'' in the ``sky''. This multi-label nature makes image classification intrinsically different from simple binary classification.

Moreover, different labels cannot be properly characterized by a single feature representation. For example, the color information (e.g. color histogram), shape cue (encoded in SIFT \cite{Lowe-IJCV-2004}) and global structure (e.g. GIST \cite{Oliva-and-Torralba-IJCV-2001}) can effectively represent natural substances (e.g. sky, cloud and plant life), man-made objects (e.g. aeroplane, motorbike and TV-monitor) and scenes (e.g. seaside and indoor), respectively, but cannot simultaneously illustrate all these concepts in an effective way. Each visual feature encodes a particular property of images and characterizes a particular concept (label), so we treat each feature representation as a particular view for characterizing images. Fig. \ref{fig:Example_Images} (a)-(c) indicate that SIFT representation is effective in describing a motorbike and GIST can capture the global structure of a person on the motorbike. Fig. \ref{fig:Example_Images} (d)-(f) shows that GIST performs well in recognizing seaside scenes, while the color information can be used as a complementary aid for recognizing the blue sea water. From Fig. \ref{fig:Example_Images} (g)-(i) we can see that RGB usually represents cloud well and GIST is helpful when RGB fails. For example, the RGB representations of C1 and C3 are not very similar and their GIST distance (0.22) is very small due to the sky scene structure. This multi-view nature distinguishes image classification from single-view tasks, such as texture segmentation \cite{Cesmeli-and-Wang-TNN-2001} and face recognition \cite{Masip-and-Vitria-TNN-2008}.

The vector-valued function \cite{Micchelli-and-Pontil-NC-2005} has recently been introduced to resolve multi-label classification \cite{Minh-and-Sindhwani-ICML-2011} and has been demonstrated to be effective in semantic scene annotation. This method naturally incorporates the label-dependencies into the classification model by first computing the graph Laplacian \cite{Belkin-et-al-JMLR-2006} of the output similarity graph, and then using this graph to construct a vector-valued kernel. This model is superior to most of the existing multi-label learning methods \cite{Chen-et-al-SDM-2008, Hariharan-et-al-ICML-2010, Sun-et-al-PAMI-2011} because it naturally considers the label correlations and efficiently outputs all the predicted labels at one time.

Although the vector-valued function is effective for general multi-label classification tasks, it cannot directly handle image classification problems that include images represented by multi-view features. A popular solution is to concatenate all the features into a long vector. This concatenation strategy not only ignores the physical interpretations of different features, however, but also encounters the over-fitting problem given limited training samples.

We thus introduce multi-kernel learning (MKL) to the vector-valued function and present a multi-view vector-valued manifold regularization (MV$^3$MR) framework for handling the multi-view features in multi-label image classification. MV$^3$MR associates each view with a particular kernel, assigns a higher weight to the view/kernel carrying more discriminative information, and explores the complementary nature of different views.

In particular, MV$^3$MR assembles the multi-view information through a large number of unlabeled images to discover the intrinsic geometry embedded in the high dimensional ambient space of the compact support of the marginal distribution. The local geometry, approximated by the adjacency graphs induced from multiple kernels of all the corresponding views, is more reliable than that approximated by the adjacency graph induced from a particular kernel of any corresponding view. In this way, MV$^3$MR essentially improves the vector-valued function for multi-label image classification.

Because the hinge loss is more suitable for classification than the least squares loss \cite{Rosasco-et-al-NC-2004}, we derive an SVM (support vector machine) formulation of MV$^3$MR which results in a multi-view vector-valued Laplacian SVM (MV$^3$LSVM). We carefully design the MV$^3$LSVM algorithm so that it determines the set of kernel weights in the learning process of the vector-valued function.

We thoroughly evaluate the proposed MV$^3$LSVM algorithm on two challenging datasets, PASCAL VOC' 07 \cite{Pascal-VOC-2007} and MIR Flickr \cite{MIR-Flickr-2008}, by comparing it with a popular MKL algorithm \cite{Rakotomamonjy-et-al-JMLR-2008}, a recently proposed MKL method \cite{Kloft-et-al-JMLR-2011}, and competitive multi-label learning algorithms for image classification, such as multi-label compressed sensing \cite{Hsu-et-al-NIPS-2009}, canonical correlation analysis \cite{Sun-et-al-PAMI-2011} and vector-valued manifold regularization \cite{Minh-and-Sindhwani-ICML-2011} in terms of mean average precision (mAP), mean area under curve (mAUC) and hamming loss (HL). The experimental results suggest the effectiveness of MV$^3$LSVM.

The rest of the paper is organized as follows. Section \ref{sec:Related_Work} summarizes the recent work in multi-label learning, multi-kernel learning and image classification. In Section \ref{sec:MR_and_VVMR}, we introduce manifold regularization and its vector-valued generalization. We depict the proposed MV$^3$MR framework and its SVM formulation in Section \ref{sec:MV3MR}. Extensive experiments are presented in Section \ref{sec:Experiment} and we conclude this paper in Section \ref{sec:Conclusion}.


\section{Related Work}
\label{sec:Related_Work}

\subsection{Multi-label learning}
Multi-label classification has received intensive attention in recent years. Some methods extend traditional multi-class algorithms to cope with the multi-label problem. AdaBoost.MH \cite{Schapire-and-Singer-ML-2000} adds the label value to the feature vector and then applies AdaBoost on weak classifiers. A ranking algorithm is presented in \cite{Elisseeff-and-Weston-NIPS-2001} by adopting the ranking loss as the cost function in SVM. ML-KNN \cite{Zhang-and-Zhou-PR-2007} is an extension of the $k$-nearest neighbor (KNN) algorithm to deal with multi-label data and canonical correlation analysis (CCA) has also recently been extended to the multi-label case by formulating it as a least-squares problem \cite{Sun-et-al-PAMI-2011}.

Other works concentrate on preprocessing the data so that standard binary or multi-class techniques can be utilized. For example, multiple labels of a sample belong to a subset of the whole label set and we can view this subset as a new class \cite{McCallum-AAAI-1999}. This may lead to a large number of classes and a more common strategy is to learn a binary classifier for each label \cite{Boutell-et-al-PR-2004, Guillaumin-et-al-CVPR-2010}. Considering that the labels are often sparse, a compressed sensing method is proposed for multi-label prediction \cite{Hsu-et-al-NIPS-2009}.

Various approaches have been proposed to improve prediction accuracy by exploiting label correlations \cite{Sun-et-al-KDD-2008, Ji-et-al-TKDD-2010, Hariharan-et-al-ICML-2010, Minh-and-Sindhwani-ICML-2011}. Sun et al. \cite{Sun-et-al-KDD-2008} proposed the construction of a hypergraph to exploit the label dependencies. In \cite{Ji-et-al-TKDD-2010}, a common subspace is assumed to be shared among all labels, and the correlation information contained in different labels can be captured by learning this low-dimensional subspace. A max-margin method is proposed in \cite{Hariharan-et-al-ICML-2010}, where the prior knowledge of the label correlations is incorporated explicitly in the multi-label classification model.

None of the approaches mentioned above consider the features to be used; however, an image with multiple labels usually indicates that it contains multiple objects. As far as we know, there is no single kind of feature that can describe a variety of objects very well. Therefore, how to combine different features is a critical issue in multi-label image classification and we consider MKL for this purpose in this paper.



\subsection{MKL: Multi-kernel learning}
Classical kernel methods are usually based on a single kernel \cite{S-Zafeiriou-et-al-TNNLS-2012}. MKL \cite{Lanckriet-et-al-ICML-2002}, in which a kernel-based classifier and a convex combination of the kernels are learned simultaneously, has attracted much attention. Lanckriet et al. \cite{Lanckriet-et-al-ICML-2002} introduces MKL for binary classification and solves it with semi-definite programming (SDP) techniques. The MKL problem is further developed by Sonnenburg et al. \cite{Sonnenburg-et-al-JMLR-2006} in the presentation of a semi-infinite linear program (SILP). In \cite{Rakotomamonjy-et-al-JMLR-2008}, MKL is reformulated by using a weighted L2-norm regularization to replace the mixed-norm regularization and adding an L1-norm constraint on the kernel weights. All of these MKL formulations are based on SVM and are not naturally designed for multi-label classification. The proposed MV$^3$MR framework extends MKL to handle the multi-label problem and model label inter-dependencies.

\subsection{Image classification}
Image classification has been widely used in many computer vision-related applications such as image retrieval and web content browsing. In recent years, more than a dozen methods have been proposed and representative works can be grouped into three categories.\\
$\bullet$ \textbf{Single-view learning for image classification:} This category contains many recent image classification schemes, e.g. dictionary learning \cite{Kreutz-et-al-NC-2003} and spatial pyramid matching \cite{Lazebnik-et-al-CVPR-2006}. For example, Labusch et al. \cite{Labusch-et-al-TNN-2008} proposed to integrate sparse-coding and local maximum operation to extract local features for handwritten digit recognition. In \cite{Zhou-et-al-ECCV-2010}, a non-linear coding scheme was introduced for local descriptors such as SIFT. Yang et al. \cite{Yang-and-Newsam-ICCV-2011} explored the local co-occurrences of visual words over the spatial pyramid.\\
$\bullet$ \textbf{Multi-view learning for image classification:} Schemes in this category utilize the features from different views (or multi-view features) to boost image classification performance. In this paper, the concept ``views'' used for learning refer to different features or attributes for depicting the objects to be classified. It should be noted that for some other applications in vision and graphics, the ``views'' mean different spatial viewpoints \cite{SuH-et-al-ICCV-2009, FuY-et-al-TMM-2010, A-Iosifidis-et-al-TNNLS-2012}. A semi-supervised boosting algorithm is proposed in \cite{Saffari-et-al-ECCV-2010}, in which images measured by different views are used to construct a prior and formulate a regularization term. Guillaumin et al. \cite{Guillaumin-et-al-CVPR-2010} combined 15 visual representations (e.g. SIFT, GIST and HSV) with the tag feature for semi-supervised image classification. Combining the visual and textual information has been utilized for clustering \cite{D-Zhang-et-al-TNN-2011} and web page classification \cite{H-Zhang-et-al-TNN-2011}.\\
$\bullet$ \textbf{Multi-label learning for image classification:} This category is motivated by the success of multi-label learning and has demonstrated promising image classification performance. For example, Bucak et al. \cite{Bucak-et-al-CVPR-2011} proposed a ranking based algorithm to tackle the multi-label problem with incompletely labeled data by introducing a group lasso regularizer in optimization. Unlike traditional multi-label methods that always consider positive label correlations, a novel approach is presented in \cite{Chen-et-al-ICCV-2011} to make use of the negative relationship of categories.

Although it has been widely acknowledged that both multi-view representation and label inter-dependencies are important for multi-label image classification, most of the existing approaches do not take both of them into consideration. Most existing multi-view approaches assume that different views (features) contribute equally to label prediction. In contrast to these approaches, the proposed MV$^3$MR naturally explores both the complementary property of multi-view features and the correlations of different labels under the manifold regularization scheme.

\section{Manifold Regularization and Vector-valued Generalization}
\label{sec:MR_and_VVMR}

This section briefly introduces the manifold regularization framework \cite{Belkin-et-al-JMLR-2006} and its vector-valued generalization \cite{Minh-and-Sindhwani-ICML-2011}. Given a set of $l$ labeled examples $D_l=\{(x_i,y_i)_{i=1}^l\}$ and a relatively large set of $u$ unlabeled examples $D_u=\{(x_i)_{i=l+1}^{N=l+u}\}$, we consider a non-parametric estimation of a vector-valued function $f:\mathcal{X} \mapsto \mathcal{Y}$, where $\mathcal{Y}=\mathbb{R}^n$ and $n$ is the number of labels. This setting includes $\mathcal{Y}=\mathbb{R}$ as a special case for regression and classification.

\subsection{Manifold regularization}
Manifold learning has been widely used for capturing the local geometry \cite{Z-Fan-et-al-TNN-2011} and conducting low-dimensional embedding \cite{D-Bouchaffra-TNNLS-2012, L-Chen-et-al-TNNLS-2012}. In manifold regularization, the data manifold is characterized by a nearest neighbor graph $\mathcal{W}$, which explores the geometric structure of the compact support of the marginal distribution. The Laplacian $\mathcal{L}$ of $\mathcal{W}$ and the prediction $\mathbf{f}=[f(x_1),\ldots,f(x_N)]$ are then formulated as a smoothness constraint $\|f\|_I^2 = \mathbf{f}^T \mathcal{L} \mathbf{f}$, where $\mathcal{L}=\mathcal{D}-\mathcal{W}$ and the diagonal matrix $\mathcal{D}$ is given by $\mathcal{D}_{ii}=\sum_{j=1}^N \mathcal{W}_{ij}$. The manifold regularization framework minimizes the regularized loss
\begin{equation}
\label{eq:MR_Formulation}
\mathop{\mathrm{argmin}}_{f \in \mathcal{H}_k} \frac{1}{l} \sum_{i=1}^l L(f,x_i,y_i) + \gamma_A \|f\|_k^2 + \gamma_I \|f\|_I^2
\end{equation}
where $L$ is a predefined loss function; $k$ is the standard scalar-valued kernel, i.e. $k:\mathcal{X} \times \mathcal{X} \mapsto \mathbb{R}$ and $\mathcal{H}_k$ is the associated reproducing kernel Hilbert space (RKHS). Here, $\gamma_A$ and $\gamma_I$ are trade-off parameters to control the complexities of $f$ in the ambient space and the compact support of the marginal distribution. The Representer theorem \cite{Belkin-et-al-JMLR-2006} ensures the solution of problem (\ref{eq:MR_Formulation}) takes the form $f^*(x)=\sum_{i=1}^N \alpha_i k(x,x_i)$, where $\alpha_i \in \mathbb{R}$ is the coefficient. Since a pair of close samples means that the corresponding conditional distributions are similar, the manifold regularization $\|f\|_I^2$ helps the function learning.

\subsection{Vector-valued manifold regularization}
In the vector-valued RKHS, where a kernel function $K$ is defined and the corresponding $\mathcal{Y}$-valued RKHS is denoted by $\mathcal{H}_K$, the optimization problem of the vector-valued manifold regularization (VVMR) is given by
\begin{equation}
\label{eq:VVMR_Formulation}
\mathop{\mathrm{argmin}}_{f \in \mathcal{H}_K} \frac{1}{l} \sum_{i=1}^l L(f,x_i,y_i) + \gamma_A \|f\|_K^2 + \gamma_I \langle \mathbf{f},\mathcal{M}\mathbf{f} \rangle_{\mathcal{Y}^{u+l}}
\end{equation}
where $\mathcal{Y}^{u+l}$ is the $u+l$-direct product of $\mathcal{Y}$ and the inner product takes the form
\begin{equation}
\notag
\langle (y_1,\ldots,y_{u+l}), (w_1,\ldots,w_{u+l}) \rangle_{\mathcal{Y}^{u+l}} = \sum_{i=1}^{u+l} \langle y_i,w_i \rangle_\mathcal{Y}
\end{equation}
The function prediction $\mathbf{f}=[f(x_1),\ldots,f(x_{u+l})] \in \mathcal{Y}^{u+l}$. The matrix $\mathcal{M}$ is a symmetric positive operator that satisfies $\langle \mathbf{y},\mathcal{M}\mathbf{y} \rangle \geq 0$ for all $\mathbf{y} \in \mathcal{Y}^{u+l}$ and is chosen to be $\mathcal{L} \otimes I_n$. Here, $\mathcal{L}$ is the graph Laplacian, $I_n$ is the $n \times n$ identity matrix and $\otimes$ denotes the Kronecker (tensor) matrix product. For $\mathcal{Y}=\mathbb{R}^n$, an entry $K(x_i,x_j)$ of the $n \times n$ vector-valued kernel matrix is defined by
\begin{equation}
\label{eq:Vector_Valued_Kernel}
K(x_i,x_j) = k(x_i,x_j) (\gamma_O L_{out}^\dag + (1-\gamma_O) I_n)
\end{equation}
where $k(\cdot,\cdot)$ is a scalar-valued kernel, $\gamma_O \in [0,1]$ is a parameter. Here, $\mathcal{L}_{out}^\dag$ is the pseudo-inverse of the output labels' graph Laplacian. The graph can be estimated by viewing each label as a vertex and using the nearest neighbors method. The representation of the $j$'th label is the $j$'th column in the label matrix $Y \in \mathbb{R}^{N \times n}$, in which, $Y_{ij}=1$ if the $j$'th label is manually assigned to the $i$'th sample, and $-1$ otherwise. For the unlabeled samples, $Y_{ij}=0$.

It has been proved in \cite{Minh-and-Sindhwani-ICML-2011} that the solution of the minimization problem (\ref{eq:VVMR_Formulation}) takes the form $f^*(x)=\sum_{i=1}^N K(x,x_i) a_i$. By choosing the Regularization Least Squares (RLS) loss $L(f,x_i,y_i)=(f(x_i)-y_i)^2$, we can estimate the column vector $\mathbf{a}=\{a_1,\ldots,a_{u+l}\} \in \mathbb{R}^{n(u+l)}$ with each $a_i \in \mathcal{Y}$ by solving a \emph{Sylvester Equation}:
\begin{equation}
\label{eq:VVLRLS_Solution}
-\frac{1}{l \gamma_A} (J_l^N G^k + l \gamma_I \mathcal{L} G^k)AQ - A + \frac{1}{l \gamma_A} Y = 0,
\end{equation}
where $a=vec(A^T)$ and $Q=(\gamma_O \mathcal{L}_{out}^\dag + (1-\gamma_O) I_n)$ and $J_l^N$ is a diagonal matrix with the first $l$ entries $1$, and the others $0$. Here, $G^k$ is the Gram matrix of the scalar-valued kernel $k$ over the labeled and unlabeled data. We refer to \cite{Minh-and-Sindhwani-ICML-2011} for a detailed description of the vector-valued Laplacian RLS.

\begin{figure}[!t]
\centering
\includegraphics[width=1.0\columnwidth]{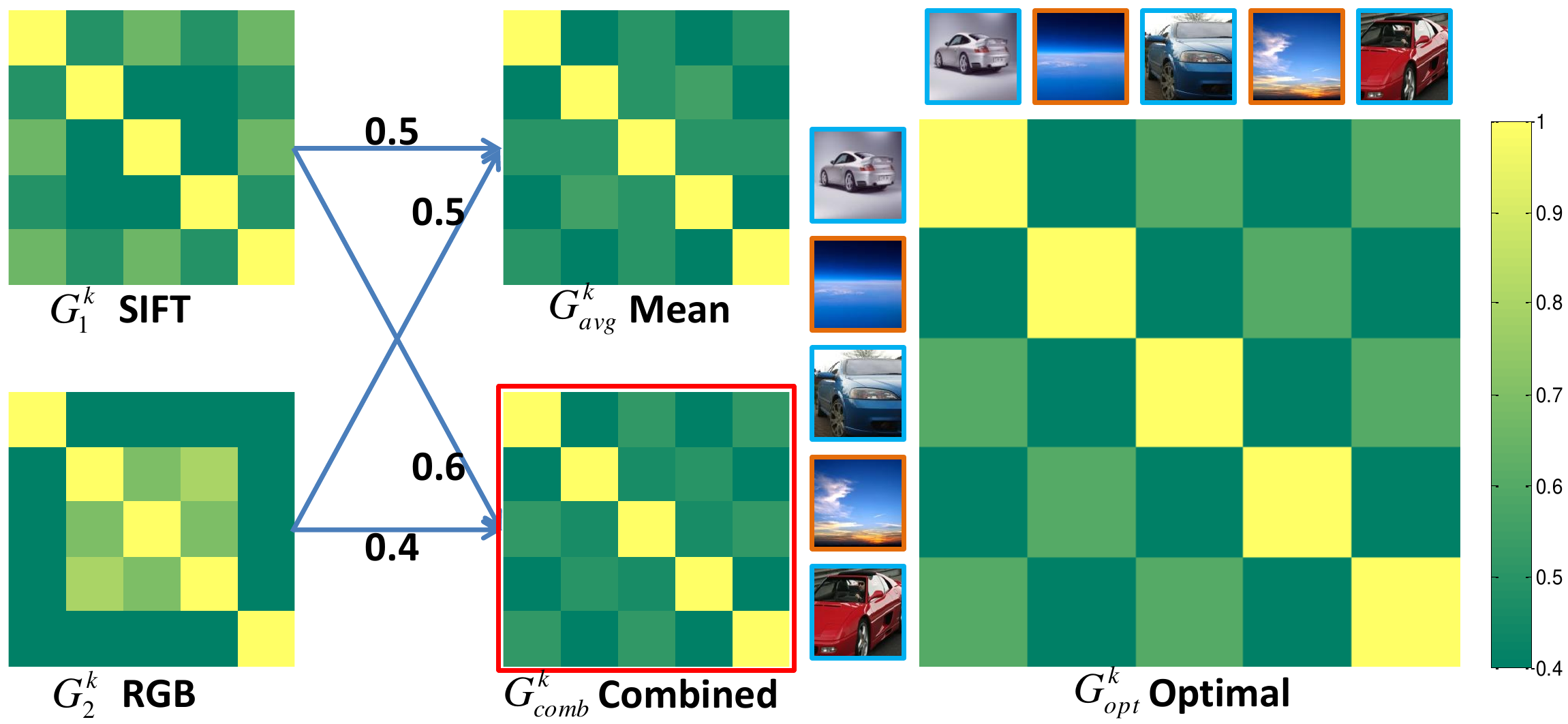}
\caption{Different views contribute to the classification differently. $G_1^k$ is a Gram matrix constructed from SIFT \cite{Lowe-IJCV-2004}. $G_2^k$ is obtained from RGB color histogram. $G_1^k$ and $G_2^k$ are complementary to each other. The learned linear combination $G_{comb}^k$ of the two Gram matrices is closer to the optimal Gram matrix $G_{opt}^k$ than the mean Gram matrix $G_{avg}^k$ by simply averaging the two kernels.}
\label{fig:Multiview_Motivation}
\vspace{-5mm}
\end{figure}

\section{MV$^3$MR: Multi-view Vector-valued Manifold Regularization}
\label{sec:MV3MR}

To handle multi-view multi-label image classification, we generalize VVMR and present multi-view vector-valued manifold regularization (MV$^3$MR). In contrast to \cite{Guillaumin-et-al-CVPR-2010}, which assumes that different views contribute equally to the classification, MV$^3$MR assumes that different views contribute to the classification differently and learns the combination coefficients to integrate different views.

Fig. \ref{fig:Multiview_Motivation} gives an illustrative example which suggests that different views contribute to the classification differently, and that learning the combination coefficients to integrate different views benefits the classification. Given five images from two classes, namely three cars of different colors (silvery white, blue and red) and two different sky images, the optimal Gram matrix $G_{opt}^k$ is shown on the right side for separating these images into two classes. On the left, there are four Gram matrices, which are two single Gram matrices $G_1^k$, $G_2^k$ obtained from two different views, and their mean $G_{avg}^k$, as well as their linear combination $G_{comb}^k$ with the learned coefficients. The figure indicates that $G_{comb}^k$ is closer to the optimal Gram matrix $G_{opt}^k$ than $G_{avg}^k$.

Given a small number of labeled samples and a relative large number of unlabeled samples, MV$^3$MR first computes an output similarity graph by using the label information of the labeled samples. The Laplacian of the label graph is incorporated in the scalar-valued Gram matrix $G_v^k$ over labeled and unlabeled data to enforce label correlations on each view, and the vector-valued Gram matrices $G_v=G_v^k \otimes Q,v=1,\ldots,V$ can be obtained. Meanwhile, we also compute the vector-valued graph Laplacians $\mathcal{M}_v,v=1,\ldots,V$ by using the features of the input data from different views. Then MV$^3$MR learns the kernel combination coefficient $\beta_v$ for $G_v$ as well as the graph weight $\theta_v$ for $\mathcal{M}_v$ by the use of alternating optimization. Finally, the combined Gram matrix $G$ together with the regularization on the combined manifold $M$ is used for classification. Fig. \ref{fig:System_Diagram} summarizes the above procedure. Technical details are given below.

\begin{figure}[!t]
\centering
\includegraphics[width=1.0\columnwidth]{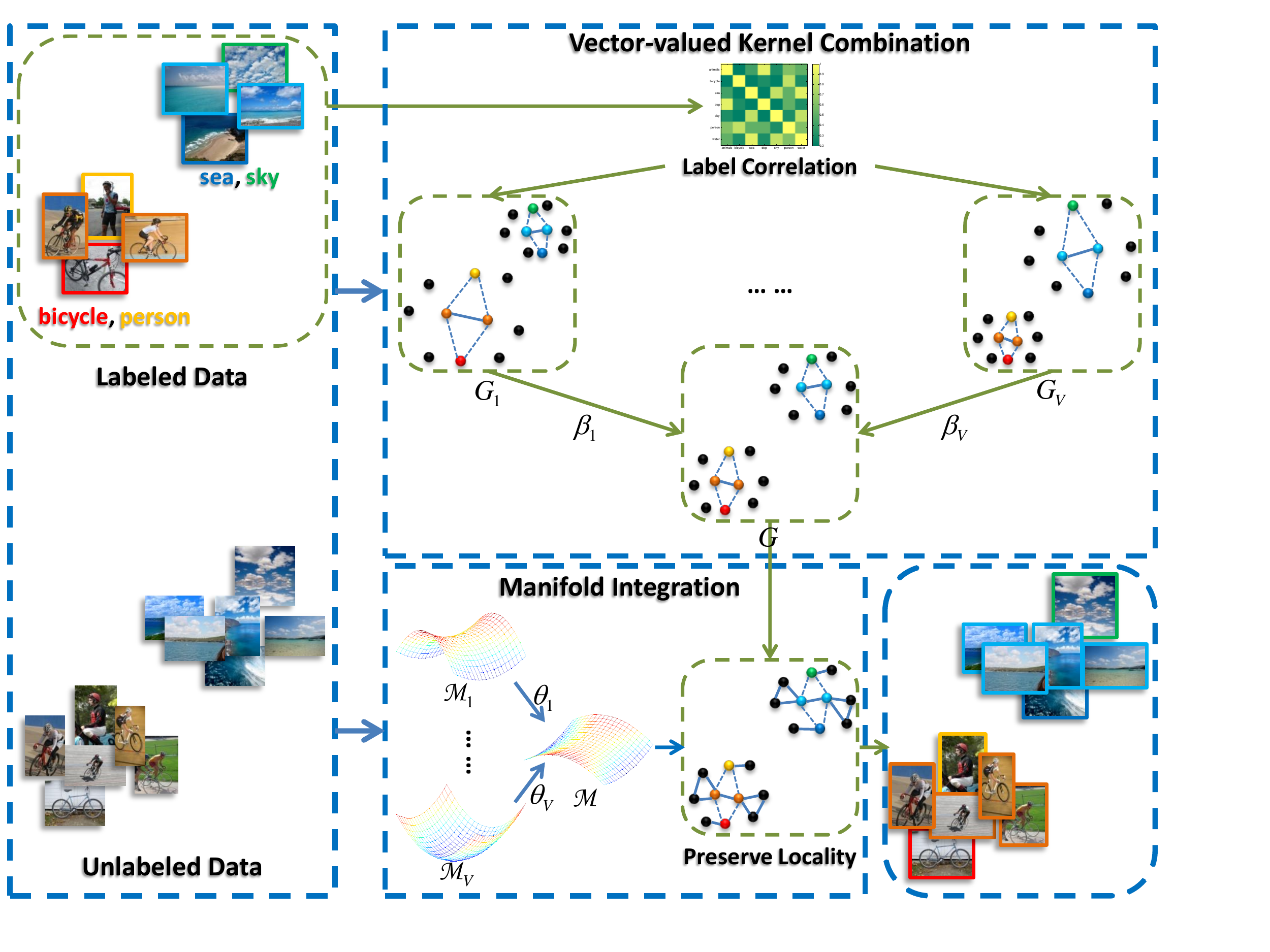}
\caption{The diagram of the proposed MV$^3$MR algorithm. The given labels are used to construct an output similarity graph, which encodes the label correlations. Features from different views of the labeled and unlabeled data are used to construct different Gram matrices (with label correlations incorporated) $G_v,v=1,\ldots,V$ as well as the different graph Laplacians $\mathcal{M}_v,v=1,\ldots,V$. We learn the weight $\beta_v$ for $G_v$ and $\theta_v$ for $\mathcal{M}_v$ respectively. The combined Gram matrix $G$ is used for classification while preserving locality on the integrated manifold $\mathcal{M}$.}
\label{fig:System_Diagram}
\vspace{-5mm}
\end{figure}

\subsection{Rationality}
\label{subsec:Rationality}
Let $V$ be the number of views and $v$ be the view index. On the feature space of each view, we define the corresponding positive definite scalar-valued kernel $k_v$, which is associated with an RKHS $\mathcal{H}_{k_v}$. It follows from the functional framework \cite{Rakotomamonjy-et-al-JMLR-2008} that by introducing a non-negative coefficient $\beta_v$, the Hilbert space $\mathcal{H}_{k_v}^\prime = \{ f|f \in \mathcal{H}_{k_v}:\frac{\|f\|_{\mathcal{H}_{k_v}}}{\beta_v}<\infty \}$ is an RKHS with kernel $k(x,x^\prime)=\beta_v k_v (x,x^\prime)$. If we define $\mathcal{H}_k$ as the direct sum of the space $\mathcal{H}_{k_v}^\prime$, i.e. $\mathcal{H}_k=\oplus_{v=1}^V \mathcal{H}_{k_v}^\prime$, then $\mathcal{H}_k$ is an RKHS associated with the kernel
\begin{equation}
\label{eq:Kernel_Combination}
k(x,x^\prime) = \sum_{v=1}^V \beta_v k_v(x,x^\prime)
\end{equation}
Thus, any function in $\mathcal{H}_k$ is a sum of functions belonging to $\mathcal{H}_{k_v}$. The vector-valued kernel $K(x,x^\prime) = k(x,x^\prime) \otimes Q = \sum_{v=1}^V \beta_v K_v(x,x^\prime)$, where we have used the bilinearity of the Kronecker product. Each $K_v (x,x^\prime) = k_v(x,x^\prime) \otimes Q$ corresponds to an RKHS according to the study of RKHS for the vector-valued functions \cite{Minh-and-Sindhwani-ICML-2011}. Thus, the kernel $K$ is associated with an RKHS $\mathcal{H}_K$. This functional framework motivates the MV$^3$MR framework. We will jointly learn the linear combination coefficients $\{\beta_v\}$ to integrate kernels for characterizing different views and the classifier coefficients $\{a_i\}$ in a single optimization problem. Moreover, to effectively utilize the unlabeled data, we construct graph Laplacians for different views and learn to combine all of them.

\vspace{-2mm}
\subsection{Problem formulation}
Under the multi-view setting and the theme of manifold regularization, we propose to learn the vector-valued function $f$ by linearly combining the kernels and graphs from different views. The optimization problem is given by
\begin{equation}
\label{eq:MV3MR_Formulation}
\begin{split}
\mathop{\mathrm{argmin}}_{f \in \mathcal{H}_K}
& \frac{1}{l} \sum_{i=1}^l L(f,x_i,y_i) + \gamma_A \|f\|_K^2 + \gamma_I \langle \mathbf{f},\mathcal{M}\mathbf{f} \rangle_{\mathcal{Y}^{u+l}} \\
& + \gamma_B \|\beta\|_2^2 + \gamma_C \|\theta\|_2^2 \\
\mathrm{s.t.} \
& \sum_{v=1}^V \beta_v = 1, \beta_v \geq 0, \\
& \sum_{v=1}^V \theta_v = 1, \theta_v \geq 0, v = 1,\ldots,V,
\end{split}
\end{equation}
where $\beta=[\beta_1,\ldots,\beta_V]^T$ and $\theta=[\theta_1,\ldots,\theta_V]^T$. Both $\gamma_B>0$ and $\gamma_C>0$ are trade-off parameters. The decision function takes the form $f(x)+b=\sum_v f^v(x)+b$ and belongs to an RKHS $\mathcal{H}_K$ associated with the kernel $K(x,x^\prime)=\sum_v \beta_v K_v (x,x^\prime)$. We define $\mathcal{M}=\sum_v \theta_v \mathcal{M}_v$, where each $\mathcal{M}_v$ is a vector-valued graph Laplacian constructed on $\mathcal{H}_{K_v}$. It can be demonstrated that $\mathcal{M}$ is still a graph Laplacian.
\newtheorem{lemma}{Lemma}
\begin{lemma}
\label{lm:M_Graph_Laplacian}
$\mathcal{M} \in S_{Nn}^+$ is a vector-valued graph Laplacian.
\end{lemma}
The notation $S_n^+$ denotes a set of $n \times n$ symmetric positive semi-definite matrices and we will use $S_n^*$ to denote a set of positive definite matrices. Then we have the following version of the Representer Theorem.
\newtheorem{theorem}{Theorem}
\begin{theorem}
\label{thm:Representer_Theorem}
For fixed sets of $\{\beta_v\}$ and $\{\theta_v\}$, the minimizer of problem (\ref{eq:MV3MR_Formulation}) admits an expansion
\begin{equation}
\label{eq:MV3MR_Predictive_Function}
f^*(x) = \sum_{i=1}^{u+l} K(x,x_i) a_i,
\end{equation}
\end{theorem}
where $a_i \in \mathcal{Y}, 1 \leq i \leq N=u+l$ are some vectors to be estimated and $K(x,x_i)=\sum_{v=1}^V \beta_v K_v(x,x_i)$. The proof of Lemma \ref{lm:M_Graph_Laplacian} and Theorem \ref{thm:Representer_Theorem} are detailed in the appendix.

The hinge loss $L(f,x_i,y_i)=(1-y_i f(x_i ))_+$ is more suitable for classification than least squares loss since the hinge loss results in a better convergence rate and usually higher classification accuracy; we refer to \cite{Rosasco-et-al-NC-2004} for a comparison of different popular loss functions. We adopt the hinge loss in MV$^3$MR and derive MV$^3$LSVM as follows.

\subsection{Multi-view vector-valued Laplacian SVM}
Under the SVM formulation, the minimization problem of MV$^3$MR is
\vspace{-5mm}
\begin{equation}
\label{eq:MV3LSVM_Formulation}
\begin{split}
\mathop{\mathrm{argmin}}_{f \in \mathcal{H}_K, \beta, \theta}
& \frac{1}{nl} \sum_{i=1}^l \sum_{j=1}^n (1-y_{ij}f_j(x_i))_+ + \gamma_A \|f\|_K^2 \\
& + \gamma_I \langle \mathbf{f},\mathcal{M}\mathbf{f} \rangle_{\mathcal{Y}^{u+l}} + \gamma_B \|\beta\|_2^2 + \gamma_C \|\theta\|_2^2 \\
\mathrm{s.t.} \
& \sum_{v=1}^V \beta_v = 1, \beta_v \geq 0, \mathrm{and} \sum_{v=1}^V \theta_v = 1, \theta_v \geq 0, \forall v,
\end{split}
\end{equation}
An unregularized bias $b_j$ is often added to the solution $f_j(x)=\sum_{i=1}^N K_j(x,x_i)a_i$ in the SVM formulation. By substituting (\ref{eq:MV3MR_Predictive_Function}) into the above formulation, we can see the primal problem as follows:
\begin{equation}
\label{eq:MV3LSVM_Primal_Problem}
\begin{split}
\mathop{\mathrm{argmin}}_{\mathbf{a}, b, \mathbf{\xi}, \beta, \theta}
& \frac{1}{nl} \sum_{i=1}^l \sum_{j=1}^n \xi_{ij} + \gamma_A \mathbf{a}^T G \mathbf{a} + \gamma_I \mathbf{a}^T G \mathcal{M} G \mathbf{a} \\
& + \gamma_B \|\beta\|_2^2 + \gamma_C \|\theta\|_2^2 \\
\mathrm{s.t.} \
& y_{ij}(\sum_{z=1}^{l+u} K_j(x_i,x_z)a_z+b_j) \geq 1-\xi_{ij}, \xi_{ij} \geq 0, \forall i,j \\
& \sum_{v=1}^V \beta_v = 1, \beta_v \geq 0, \mathrm{and} \sum_{v=1}^V \theta_v = 1, \theta_v \geq 0, \forall v,
\end{split}
\end{equation}
where $G=\sum_{v=1}^V \beta_v G_v$ is the combined vector-valued Gram matrix over the labeled and unlabeled samples defined on kernel $K$, $\mathcal{M}=\sum_{v=1}^V \theta_v \mathcal{M}_v$ is the integrated vector-valued graph Laplacian. Here, $K_j(\cdot,\cdot)$ is the $j$th row of the vector-valued kernel $K$. We have three variables, i.e., $\mathbf{a}$, $\beta$ and $\theta$, to be optimized in (\ref{eq:MV3LSVM_Primal_Problem}). To solve this problem, we consider the following constrained optimization problem:
\begin{equation}
\label{eq:MV3LSVM_Contrained_Optimization}
\begin{split}
& \mathrm{min} F(\beta, \theta) \\
& \mathrm{s.t.} \sum_{v=1}^V \beta_v=1, \beta_v \geq 0, \mathrm{and} \sum_{v=1}^V \theta_v=1, \theta_v \geq 0, \forall v.
\end{split}
\end{equation}
where $F(\beta, \theta)$ equals to
\begin{numcases}
{}{}
\label{eq:MV3LSVM_Primal_Problem_A}
\begin{split}
\mathop{\mathrm{argmin}}_{\mathbf{a}, b, \mathbf{\xi}}
& \frac{1}{nl} \sum_{i=1}^l \sum_{j=1}^n \xi_{ij} + \gamma_A \mathbf{a}^T G \mathbf{a} + \gamma_I \mathbf{a}^T G \mathcal{M} G \mathbf{a} \\ & + \gamma_B \|\beta\|_2^2 + \gamma_C \|\theta\|_2^2 \\
\mathrm{s.t.} \
& y_{ij}(\sum_{z=1}^{l+u} K_j(x_i,x_z)a_z+b_j) \geq 1-\xi_{ij}, \\
& \xi_{ij} \geq 0, i=1,\ldots,l, j=1,\ldots,n,
\end{split}
\end{numcases}
Here, $G$ and $\mathcal{M}$ take the form as in (\ref{eq:MV3LSVM_Primal_Problem}). We can omit the terms $\gamma_B \|\beta\|_2^2$ and $\gamma_C \|\theta\|_2^2$ in (\ref{eq:MV3LSVM_Primal_Problem_A}) since ¦Â and ¦È are fixed.  By introducing the Lagrange multipliers $\mu_{ij}$  and $\eta_{ij}$ in (\ref{eq:MV3LSVM_Primal_Problem_A}), we have
\begin{equation}
\label{eq:MV3LSVM_Lagrange_Multiplier_Formulation_A}
\begin{split}
& W(\mathbf{a},\xi,b,\mu,\eta) \\
& = \frac{1}{nl} \sum_{i=1}^l \sum_{j=1}^n \xi_{ij} + \frac{1}{2} \mathbf{a}^T (2 \gamma_A G + 2 \gamma_I G \mathcal{M} G) \mathbf{a} - \sum_{i=1}^l \sum_{j=1}^n \eta_{ij} \xi_{ij} \\
& - \sum_{i=1}^l \sum_{j=1}^n \mu_{ij} \Big(y_{ij} \big(\sum_{z=1}^{l+u} K_j(x_i,x_z)a_z+b_j\big)-1+\xi_{ij}\Big). \\
\end{split}
\end{equation}
By taking the partial derivative w.r.t. $\xi_{ij}$, $b_j$, and setting them to be zero, we obtain
\begin{equation}
\notag
\begin{split}
& \frac{\partial W}{\partial b_j}=0 \Rightarrow \sum_{i=1}^l \mu_{ij} y_{ij}=0, j=1,\ldots,n, \\
& \frac{\partial W}{\partial \xi_{ij}}=0 \Rightarrow \frac{1}{nl}-\mu_{ij}-\eta_{ij}=0 \Rightarrow 0 \leq \mu_{ij} \leq \frac{1}{nl}.
\end{split}
\end{equation}
A reduced Lagrangian can be obtained by substituting the above equalities back into (\ref{eq:MV3LSVM_Lagrange_Multiplier_Formulation_A}), which leads to
\begin{equation}
\label{eq:MV3LSVM_Reduced_Lagrangian_A}
\begin{split}
W^R(\mathbf{a},\mu) & = \frac{1}{2}\mathbf{a}^T \Big(2\gamma_A G + 2\gamma_I G \mathcal{M} G\Big)\mathbf{a} - \mathbf{a}^T G J^T Y_d \mu + \mu^T \mathbf{1}, \\
\mathrm{s.t.} \
& \sum_{i=1}^l \mu_{ij} y_{ij} = 0, j=1,\ldots,n, \\
& 0 \leq \mu_{ij} \leq \frac{1}{nl}, i=1,\ldots,l, j=1,\ldots,n,
\end{split}
\end{equation}
where $J=[I\ 0] \in \mathbb{R}^{(nl) \times (nl+nu)}$ and $I$ is an $nl \times nl$ identity matrix. Here, $\mu=\{\mu_1,\ldots,\mu_l\} \in \mathbb{R}^{nl}$ is a column vector with each $\mu_i=[\mu_{i1},\ldots,\mu_{in}]^T$, $Y_d=\mathrm{diag}(y_{11},\ldots,y_{1n},\ldots,\ldots,y_{l1},\ldots,y_{ln})$ and $\mathbf{1}$ is an all ones column vector. Taking the partial derivative of $W^R$ w.r.t. $\mathbf{a}$ and letting it be zero leads to:
\begin{equation}
\label{eq:Optimal_A}
\mathbf{a}^* = (2\gamma_A I + 2\gamma_I \mathcal{M} G)^{-1} J^T Y_d \mu^*.
\end{equation}
Substituting it back into (\ref{eq:MV3LSVM_Reduced_Lagrangian_A}) we get:
\begin{equation}
\label{eq:Optimal_Mu}
\begin{split}
\mu^* = & \mathop{argmax}_{\mu \in \mathbb{R}^{nl}} \mu^T \mathbf{1} - \frac{1}{2} \mu^T S \mu \\
\mathrm{s.t.} \
& \sum_{i=1}^l \mu_{ij} y_{ij} = 0, j=1,\ldots,n, \\
& 0 \leq \mu_{ij} \leq \frac{1}{nl}, i=1,\ldots,l, j=1,\ldots,n,
\end{split}
\end{equation}
where the matrix $S=Y_d J G(2\gamma_A I + 2\gamma_I M G)^{-1} J^T Y_d$. Again, the combined Gram matrix $G=\sum_{v=1}^V \beta_v G_v$ and the integrated graph Laplacian $\mathcal{M}=\sum_{v=1}^V \theta_v \mathcal{M}_v$. Because of the strong duality, the objective value of problem (\ref{eq:MV3LSVM_Primal_Problem_A}) is also the objective value of (\ref{eq:MV3LSVM_Reduced_Lagrangian_A}), which is $W^R(\mathbf{a}^*,\mu^*)$. Therefore, we can rewrite (\ref{eq:MV3LSVM_Contrained_Optimization}) as
\begin{equation}
\label{eq:MV3LSVM_Formulation_Beta_Theta}
\begin{split}
W(\beta,\theta) = & W^R(\mathbf{a}^*,\mu^*) + \gamma_B \|\beta\|_2^2 + \gamma_C \|\theta\|_2^2 \\
\mathrm{s.t.} \
& \sum_{v=1}^V \beta_v=1, \beta_v \geq 0; \sum_{v=1}^V \theta_v=1, \theta_v \geq 0, \forall v.
\end{split}
\end{equation}
For fixed $\theta$, the above problem can be rewritten with respect to $\beta$ as
\begin{equation}
\label{eq:MV3LSVM_Formulation_Beta}
\begin{split}
W(\beta) = & \beta^T H \beta + \gamma_B \|\beta\|_2^2 - h^T \beta \\
\mathrm{s.t.} \
& \sum_v \beta_v=1, \beta \geq 0, v=1,\ldots,V,
\end{split}
\end{equation}
where $h=[h_1,\ldots,h_V]^T$ with each $h_v=(\mathbf{a}^*)^T G_v J^T Y_d \mu^* - \gamma_A (\mathbf{a}^*)^T G_v \mathbf{a}^*$ and $H$ is a $V \times V$ matrix with the entry $H_{ij}=\gamma_I (\mathbf{a}^*)^T G_i \mathcal{M} G_j \mathbf{a}^*$. We can simply set the derivative of $W(\beta)$ to zero and obtain $\beta=(H + H^T + 2\gamma_B I)^{-1} h$. Then the computed $\beta$ is projected to the positive simplex to satisfy the summation and positive constraints. However, such an approach lacks convergence guarantees and may lead to numerical problems. A coordinate descent algorithm is therefore used to solve (\ref{eq:MV3LSVM_Formulation_Beta}). In each iteration round during the coordinate descent procedure, two elements $\beta_i$ and $\beta_j$ are selected to be updated while the others are fixed. By using the Lagrangian of problem (\ref{eq:MV3LSVM_Formulation_Beta}) and considering that $\beta_i+\beta_j$ will not change due to constraint $\sum_{v=1}^V \beta_v=1$, we have the following solution for updating $\beta_i$ and $\beta_j$
\begin{numcases}{}
\label{eq:Optimal_Beta}
\begin{split}
\beta_i^* = & \frac{2\gamma_B(\beta_i+\beta_j)+(h_i-h_j)+2t_{ij}}{2(H_{ii}-H_{ji}-H_{ij}+H_{jj})+4\gamma_B}, \\
\beta_j^* = & \beta_i+\beta_j-\beta_i^*,
\end{split}
\end{numcases}
where $t_{ij}=(H_{ii}-H_{ji}-H_{ij}+H_{jj})\beta_i - \sum_k(H_{ik}-H_{jk})\beta_k$. The obtained $\beta_i^*$ or $\beta_j^*$ may violate the constraint $\beta_v \geq 0$. Thus, we set
\begin{equation}
\notag
\begin{split}
& \beta_i^*=0, \beta_j^*=\beta_i+\beta_j, \mathrm{if}\ 2\gamma_B(\beta_i+\beta_j)+(h_i-h_j)+2t_{ij} \leq 0, \\
& \beta_j^*=0, \beta_i^*=\beta_i+\beta_j, \mathrm{if}\ 2\gamma_B(\beta_i+\beta_j)+(h_j-h_i)+2t_{ji} \leq 0.
\end{split}
\end{equation}
From the solution (\ref{eq:Optimal_Beta}), we can see that the update criteria tends to assign larger value $\beta_i$ to larger $h_i$ and smaller $H_{ii}$. Because $h_i=(\mathbf{a}^*)^T G_i J^T Y_d \mu - \gamma_A (\mathbf{a}^* )^T G_i \mathbf{a}^*$ and $H_{ii} = \gamma_I (\mathbf{a}^* )^T G_i \mathcal{M} G_i \mathbf{a}^*$ measures the discriminative ability and the performance of the $i$'th view. Let $(\mathbf{a}_i^*, \mu_i^*)$ be the solution for the optimization problem of the $i$'th view, which is $W^R(\mathbf{a}, \mu)$ with $G=G_i$. If all the solutions are the same, i.e., $(\mathbf{a}_1^*,\mu_1^*)=\ldots=(\mathbf{a}_V^*,\mu_V^*)=(\mathbf{a}^*,\mu^*)$, then the objective value $W^R(\mathbf{a}_i^*,\mu_i^*)$ of the discriminative view tends to be smaller than non-discriminative view (we assume that all Gram matrices have been normalized). A smaller $W^R(\mathbf{a}_i^*,\mu_i^*)$ corresponds to a larger $h_i$ and a smaller $H_{ii}$, and thus our algorithm prefers discriminative view. However, the solutions $(\mathbf{a}_1^*,\mu_1^*),\ldots,(a_V^*,\mu_V^*)$ may not exactly the same as $(\mathbf{a}^*,\mu^*)$. Thus the learned $\beta_i$ is in general but not strictly consistent with the performance of the $i$'th single view. We can see this in the experiments.

For fixed $\beta$, the problem (\ref{eq:MV3LSVM_Formulation_Beta_Theta}) can be simplified as
\begin{equation}
\label{eq:MV3LSVM_Formulation_Theta}
\begin{split}
W(\theta) = & s^T\theta + \gamma_C \|\theta\|_2^2 \\
\mathrm{s.t} \
& \sum_v \theta_v = 1, \theta_v \geq 0, v=1,\ldots,V,
\end{split}
\end{equation}
where $s=[s_1,\ldots,s_V]^T$ with each $s_v=\gamma_I (\mathbf{a}^*)^T G \mathcal{M}_v G \mathbf{a}^*$. Similarly, the solution of (\ref{eq:MV3LSVM_Formulation_Theta}) can be obtained by using the coordinate descent and the criteria for updating $\theta_i$ and $\theta_j$ in an iteration round is given by
\begin{numcases}{}
\label{eq:Optimal_Theta}
\begin{split}
& \theta_i^*=0, \theta_j^*=\theta_i+\theta_j, \mathrm{if}\ 2\gamma_C(\theta_i+\theta_j)+(s_j-s_i) \leq 0, \\
& \theta_j^*=0, \theta_i^*=\theta_i+\theta_j, \mathrm{if}\ 2\gamma_C(\theta_i+\theta_j)+(s_i-s_j) \leq 0, \\
& \theta_i^*=\frac{2\gamma_C(\theta_i+\theta_j)+(s_j-s_i)}{4\gamma_C}, \theta_j^*=\theta_i+\theta_j-\theta_i^*, \mathrm{else}.
\end{split}
\end{numcases}
We now summarize the learning procedure of the proposed multi-view vector-valued Laplacian SVM (MV$^3$LSVM) in Algorithm \ref{alg:MV3LSVM_Learning_Procedure}.
\begin{algorithm}[!t]
\renewcommand{\algorithmicrequire}{\textbf{Input:}}
\renewcommand{\algorithmicensure}{\textbf{Output:}}
\caption{The optimization procedure of the proposed MV$^3$LSVM algorithm}
\label{alg:MV3LSVM_Learning_Procedure}
\begin{algorithmic}[1]
\REQUIRE labeled data $D_l^v=\{(x_i^v,y_i)_{i=1}^l\}$ and unlabeled data $D_u^v=\{(x_i^v)_{i=l+1}^{N=u+l}\}$ form different
views, $v=1,\ldots,V$ is the view index
\renewcommand{\algorithmicrequire}{\textbf{Algorithm parameters:}}
\REQUIRE $\gamma_A$, $\gamma_I$, $\gamma_B$ and $\gamma_C$
\ENSURE classifier variable $\mathbf{a}$, the kernel combination coefficients $\{\beta_v\}$, and the graph Laplacian weights $\{\theta_v\}$
\STATE{Construct the Gram matrix $G_v$ and the vector-valued graph Laplacian $\mathcal{M}_v$ for each view, set $\beta_v=\theta_v=\frac{1}{V},v=1,\ldots,V$; compute $G=\sum_{v=1}^V \beta_v G_v$ and $M=\sum_{v=1}^V \theta_v \mathcal{M}_v$.}
\renewcommand{\algorithmicrepeat}{\textbf{Iterate}}
\renewcommand{\algorithmicuntil}{\textbf{Until convergence}}
\REPEAT
\STATE{Approximately solve for $\mathbf{a}$ through (\ref{eq:Optimal_A}) with fixed $G$ and $\mathcal{M}$}
\STATE{Compute $\beta$ by solving (\ref{eq:MV3LSVM_Formulation_Beta}) and update the Gram matrix $G$}
\STATE{Compute $\theta$ by solving (\ref{eq:MV3LSVM_Formulation_Theta}) and update the graph Laplacian $\mathcal{M}$}
\UNTIL
\end{algorithmic}
\end{algorithm}

The stopping criterion for terminating the algorithm can be the difference of the objective value, $W^R(\mathbf{a},\mu) + \gamma_B \|\beta\|_2^2 + \gamma_C \|\theta\|_2^2$ between two consecutive steps. Alternatively, we can stop the iterations when the variation of $\beta$ and $\theta$ are both smaller than a pre-defined threshold. Our implementation is based on the difference of the objective value, i.e., if the value $|O_k-O_{k-1}|/|O_k-O_0|$ is smaller than a predefined threshold, then the iteration stops, where $O_k$ is the objective value of the $k$th iteration step. Our implementation is based on the difference of the objective value.

\subsection{Convergence analysis}
In this section, we discuss the convergence of the proposed MV$^3$LSVM algorithm. We firstly prove the convexity of the problem (\ref{eq:MV3LSVM_Primal_Problem_A}), (\ref{eq:MV3LSVM_Formulation_Beta}) and (\ref{eq:MV3LSVM_Formulation_Theta}) as follows.
\begin{IEEEproof}
The Hessian matrix of the objective function of (\ref{eq:MV3LSVM_Primal_Problem_A}) is $H_e(\mathbf{a})=\gamma_A G + \gamma_I G \mathcal{M} G$. The Gram matrix $G \in S_n^+$ and we assume that $G$ is positive definite in this paper (to enforce this property a small ridge is added to the diagonal of $G$). The second term is positive semi-definite since $x^T G \mathcal{M} G x = z^T \mathcal{M} z \geq 0$ for any $x$ and $z=Gx$. Here, we have used the property of the graph Laplacian $\mathcal{M} \in S_{Nn}^+$. Then $H_e(\mathbf{a}) \in S_{Nn}^*$ for $\gamma_A>0$ and problem (\ref{eq:MV3LSVM_Primal_Problem_A}) is strictly convex.

For the problem (\ref{eq:MV3LSVM_Formulation_Beta}), the Hessian matrix is $H_e(\beta)=H+\gamma_B I$. The matrix H is symmetric since the element $H_{ij}=H_{ij}^T=\gamma_I \mathbf{a}^T G_j \mathcal{M} G_i \mathbf{a}=H_{ji}$. In addition, the Cholesky decomposition $\mathcal{M}=P^T P$ exists since $\mathcal{M} \in S_{Nn}^+$. Let $z_i=P G_i a$, we have $H_{ij}=\gamma_I z_i^T z_j$. Thus, $H \in S_V^+$ and $H_e(\beta) \in S_V^*$ for $\gamma_B>0$. This means that (\ref{eq:MV3LSVM_Formulation_Beta}) is also strictly convex.

Finally, it is straightforward to verify that the problem (\ref{eq:MV3LSVM_Formulation_Theta}) is strictly convex for $\gamma_C>0$. This completes the proof.
\end{IEEEproof}
Now we discuss the convergence of our algorithm. Let the objective function of problem (\ref{eq:MV3LSVM_Primal_Problem}) be $R(\mathbf{a},b,\xi,\beta,\theta)$ and the initialized value be $R(\mathbf{a}^k,b^k,\xi^k,\beta^k,\theta^k)$. Since the problem (\ref{eq:MV3LSVM_Primal_Problem_A}) is convex, we have $R(\mathbf{a}^{k+1},b^{k+1},\xi^{k+1},\beta^k,\theta^k) \leq R(\mathbf{a}^k,b^k,\xi^k,\beta^k,\theta^k)$. We suppose that problem (\ref{eq:MV3LSVM_Primal_Problem}) is exactly solved, which means that the duality gap is zero. Then $R(\mathbf{a}^{k+1},b^{k+1},\xi^{k+1},\beta^k,\theta^k)=W(\beta^k,\theta^k)$. For fixed $\theta^k$, we obtain the convex problem (\ref{eq:MV3LSVM_Formulation_Beta}), thus we have $R(\mathbf{a}^{k+1},b^{k+1},\xi^{k+1},\beta^{k+1},\theta^k) \leq R(\mathbf{a}^{k+1},b^{k+1},\xi^{k+1},\beta^k,\theta^k)$. Similarly, due to the convexity of problem (\ref{eq:MV3LSVM_Formulation_Theta}), we have $R(\mathbf{a}^{k+1},b^{k+1},\xi^{k+1},\beta^{k+1},\theta^{k+1}) \leq R(\mathbf{a}^{k+1},b^{k+1},\xi^{k+1},\beta^{k+1},\theta^k)$. Therefore, the convergence of our algorithm is guaranteed.

\subsection{Complexity analysis}
For the proposed MV$^3$LSVM, the complexity is dominated by the time cost of computing $a^*$ in each iteration, where the computation of the matrix $S$ in (\ref{eq:Optimal_Mu}) involves an inversion and several multiplications of $nN \times nN$ matrix, and the time complexity is $O(n^{2.8} N^{2.8})$ using the Strassen algorithm \cite{V-Strassen-NM-1969}. Problem (\ref{eq:Optimal_Mu}) can be solved using a standard SVM solver with the time complexity $O(n^{2.3} l^{2.3})$ according to the sequential minimal optimization (SMO) \cite{J-Platt-AKM-1999}. The computations of $\beta$ and $\theta$ are quite efficient since their dimensionality is $V$, which is usually very small (e.g., $V=7$ in our experiments). Suppose the number of iterations is $k$, then the total cost of MV$^3$LSVM is $O(k(n^{2.8} N^{2.8}+n^{2.3} l^{2.3}))$. Considering that $l<N$, thus the time cost is $O(kn^{2.8} N^{2.8})$, which is $k$ times of the case that no combination coefficients ($\beta$ and $\theta$) are learned. From the experimental results shown in Section \ref{subsec:Exp_Performance_Enhancement}, we will find that $k$ is very small since our algorithm only needs a few iterations (around five) to converge. Actually, there is a balance between the time complexity and classification accuracy. If only limited number of unlabeled samples are selected to construct the input graph Laplacians, i.e., $N=u+l$ is small. Then the time complexity can be reduced with acceptable performance sacrifice. In our experiments, we obtain satisfactory accuracy by setting $N=1000$, and the time cost is acceptable.

\begin{figure*}[!t]
\centering
\includegraphics[width=1.6\columnwidth]{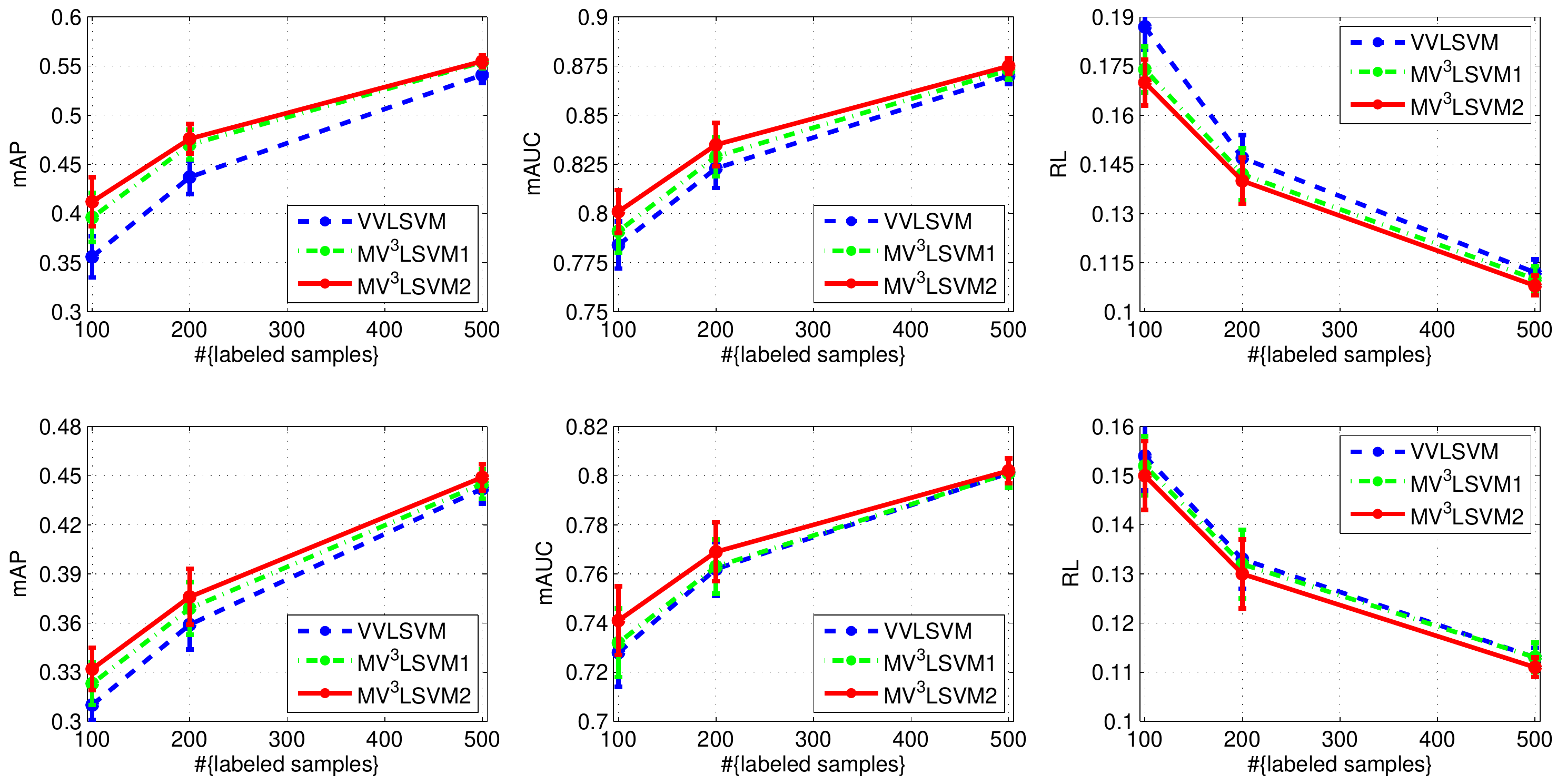}
\caption{The mAP, mAUC and RL (left to right) performance enhancement by learning the weights ($\beta$ and $\theta$) for different views: PASCAL VOC' 07 (top) and MIR Flickr (bottom). MV3LSVM1: only learn $\beta$; MV3LSVM2: learn both $\beta$ and $\theta$.}
\label{fig:Multiview_Enhancement}
\vspace{-5mm}
\end{figure*}

\section{Experiment}
\label{sec:Experiment}
We validate the effectiveness of MV$^3$LSVM on two challenge datasets, PASCAL VOC' 07 (VOC) \cite{Pascal-VOC-2007} and MIR Flickr (MIR) \cite{MIR-Flickr-2008}. The VOC dataset contains 10,000 images labeled with 20 categories. The MIR dataset consists of 25,000 images of 38 concepts. For the PASCAL VOC'07 dataset \cite{Pascal-VOC-2007}, we use the standard train/test partition \cite{Pascal-VOC-2007}, which splits 9,963 images into a training set of 5,011 images and a test set of 4,952 images. For the MIR Flickr dataset \cite{MIR-Flickr-2008}, images are randomly split into equally sized training and test sets. For both datasets, we randomly select twenty percent of the test images for validation and the rest for testing. The parameters of all the algorithms compared in our experiments are tuned by using the validation set. This means that the parameters corresponding to the best performance in the validation set are used for the transductive inference and inductive test. From the training examples, 10 random choices of $l \in \{100,200,500\}$ labeled samples are used in our experiments.

We use several visual views and the tag feature according to \cite{Guillaumin-et-al-CVPR-2010}. The visual views include SIFT features \cite{Lowe-IJCV-2004}, local hue histograms \cite{J-Weijer-and-C-Schmid-ECCV-2006}, global GIST descriptor \cite{Oliva-and-Torralba-IJCV-2001} and some color histograms (RGB, HSV and LAB). The local descriptors (SIFT and hue) are computed densely on the multi-scale grid and quantized using k-means, which will result in a visual word histogram for each image. Therefore, we have 7 different representations in total. We pre-compute a scalar-valued Gram matrix for each view and normalize it to unit trace.  For the visual representations, the kernel is defined by
\begin{equation}
\notag
k(x_i,x_j)=\mathrm{exp}\big(-\lambda^{-1} d(x_i,x_j)\big),
\end{equation}
where $d(x_i,x_j)$ denotes the distance between $x_i$ and $x_j$, $\lambda=\mathrm{max}_{i,j} d(x_i,x_j)$. Following \cite{Guillaumin-et-al-CVPR-2010}, we choose the $L1$ distance for the color histogram representations (RGB, HSV and LAB), and $L2$ for GIST and $\chi^2$ for the visual word histograms (SIFT and hue). For the tag features, a linear kernel $k(x_i,x_j)=x_i^T x_j$ is constructed.


\subsection{Evaluation metrics}
We use three kinds of evaluation criteria. The average precision (AP) and area under ROC curves (AUC) are utilized to evaluate the ranking performance under each label. We also use the ranking loss (RL) to study the performance of label set prediction for each instance.\\
$\bullet$ \emph{Average Precision (AP)} evaluates the fraction of samples ranked above a particular positive sample \cite{Zhu-Waterloo-TR-2004}. For each label, there is a ranked sequence of samples returned by the classifier. A good classifier will rank most of the positive samples higher than the negative ones. The traditional AP is defined as
\begin{equation}
\notag
\mathrm{AP} = \frac{\sum_k \mathrm{P}(k)}{\#\{\mathrm{positive\ samples}\}},
\end{equation}
where $k$ is a rank index of a positive sample and $\mathrm{P}(k)$ is the precision at the cut-off $k$. In this paper, we choose to use the computing method as in the PASCAL VOC \cite{Pascal-VOC-2007} challenge evaluation, i.e.
\begin{equation}
\notag
\mathrm{AP} = \frac{1}{11} \sum_r \mathrm{P}(r),
\end{equation}
where $\mathrm{P}(r)$ is the maximum precision over all recalls larger than $r \in \{0,0.1,0.2,\ldots,1.0\}$. A larger value means a higher performance. In this paper, the mean AP, i.e. mAP over all labels, is reported to save space.\\
$\bullet$ \emph{Area Under ROC Curves (AUC)} evaluates the probability that a positive sample will be ranked higher than a negative one by a classifier \cite{Fawcett-ML-2004}. It is computed from an ROC curve, which depicts relative trade-offs between true positive (benefits) and false positive (costs). The AUC of a realistic classifier should be larger than 0.5. We refer to \cite{Fawcett-ML-2004} for a detailed description. A larger value means a higher performance. Similar to AP, the mean AUC, i.e. mAUC over all labels, is reported.\\
$\bullet$ \emph{Ranking Loss (RL)} evaluates the fraction of label pairs that are incorrectly ranked \cite{Schapire-and-Singer-ML-2000, Zhang-and-Zhou-PR-2007}. Given a sample $x_i$ and its label set $Y_i$, a successful classifier $f(x,y)$ should have larger value for $y \in Y_i$ than those $y \not \in Y_i$. Then the ranking loss for the $i$th sample is defined as:
\begin{equation}
\notag
\begin{split}
& \mathrm{RL}(f,x_i) = \\
& \frac{1}{|Y_i|(P-|Y_i|)} |\{(y_1,y_2)|f(x,y_1) \leq f(x,y_2), y_1 \in Y_i, y_2 \not \in Y_i\}|,
\end{split}
\end{equation}
where $P$ is the total number of labels and $|\cdot|$ denotes the cardinality of a set. The smaller the value, the higher the performance. The mean value over all samples is computed for evaluation.

\begin{figure}
\centering
\includegraphics[width=0.9\columnwidth]{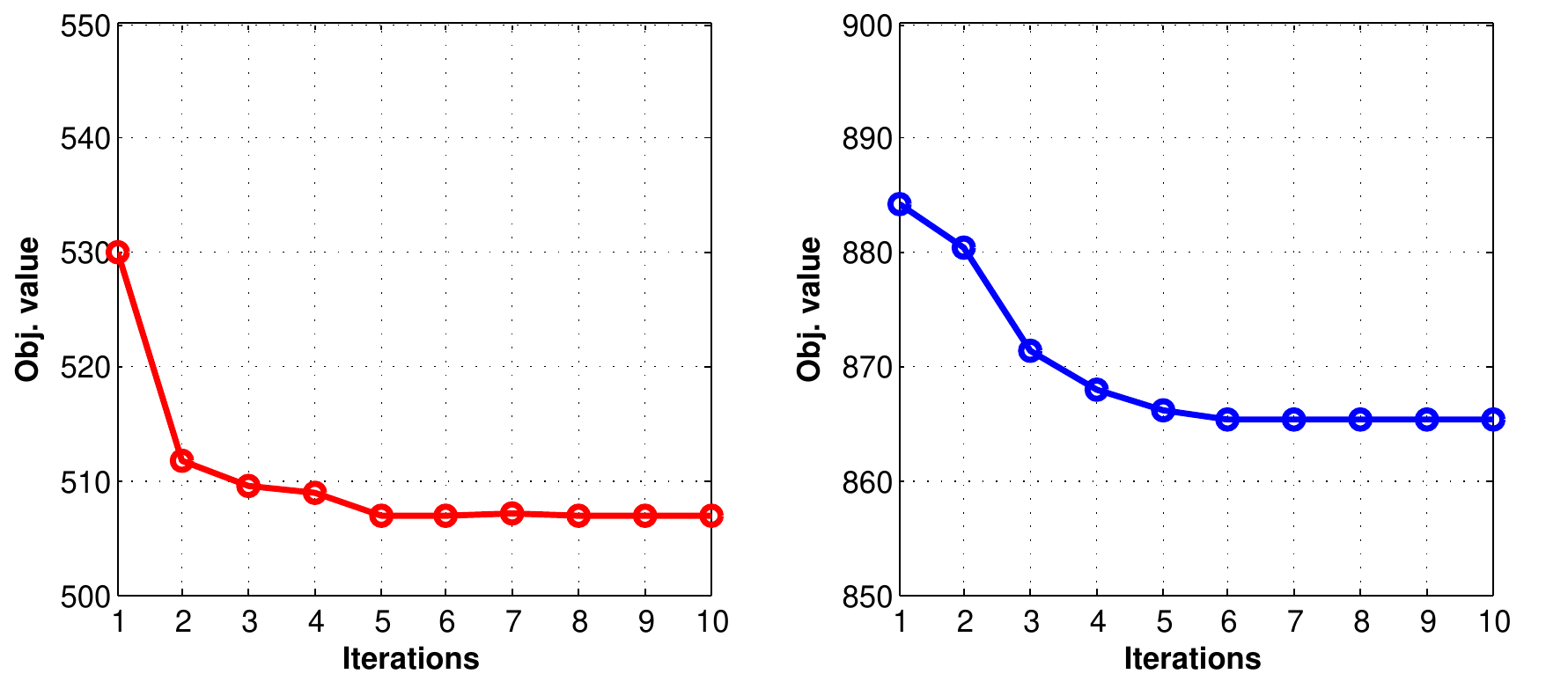}
\caption{Behavior of the objective values by increasing the iteration number on the two datasets. Left: PASCAL VOC' 07; Right: MIR Flickr.}
\label{fig:Objective_Value}
\vspace{-5mm}
\end{figure}

\subsection{Performance enhancement with multi-view learning}
\label{subsec:Exp_Performance_Enhancement}
It has been shown in \cite{Minh-and-Sindhwani-ICML-2011} that VVMR performs well for transductive semi-supervised multi-label classification and can provide a high-quality out-of-sample generalization. The proposed MV$^3$MR framework is a multi-view generalization of VVMR that incorporates the advantage from MKL for handling multi-view data. Therefore, we first evaluate the effectiveness of learning the view combination weights using the proposed multi-view learning algorithm for transductive semi-supervised multi-label classification. An out-of-sample evaluation will be presented in the next subsection. The experimental setup of the two compared methods is given as follows.\\
$\bullet$ \textbf{VVLSVM:} vector-valued Laplacian SVM, which is an SVM implementation of the vector-valued manifold regularization framework that exploits both the geometry of the input data as well as the label correlations. We do not use the vector-valued Laplacian RLS presented in \cite{Minh-and-Sindhwani-ICML-2011} for comparison because the hinge loss is more suitable for classification. The parameters $\gamma_A$ and $\gamma_I$ in (\ref{eq:VVMR_Formulation}) are both optimized over the set $\{10^i|i=-8,-7,\ldots,-2,-1\}$. We set the parameter $\gamma_O$ in (\ref{eq:Vector_Valued_Kernel}) to 1.0 since it has been demonstrated empirically in \cite{Minh-and-Sindhwani-ICML-2011} that with a larger $\gamma_O$, the performance will usually be better. The mean of the multiple Gram matrices and input graph Laplacians are pre-computed for experiments. The number of nearest neighbors for constructing the input and output graph Laplacians are tuned on the sets $\{10,20,\ldots,100\}$ and $\{2,4,\ldots,20\}$ respectively. \\
$\bullet$ \textbf{MV$^3$LSVM:} an SVM implementation of the proposed MV$^3$MR framework that combines multiple views by constructing kernels for all views and learning their weights. We tune the parameters $\gamma_A$ and $\gamma_I$ as in VVLSVM and $\gamma_O$ is set to 1.0. The additional parameters $\gamma_B$ and $\gamma_C$ are optimized over $\{10^i|i=-8,-7,\ldots,-2,-1\}$. We firstly only learn kernel combinations $\beta$ and set the graph weights $\theta$ to be uniform (MV$^3$LSVM1 in Fig. \ref{fig:Multiview_Enhancement}). Then we learn both $\beta$ and $\theta$ in MV$^3$LSVM2. We use $20$ and $6$ nearest neighbor graphs to construct the input and output normalized graph Laplacians respectively for the VOC dataset, while $30$ and $8$ nearest neighbor graphs are used in the experiments on MIR. We set these hyperparameters to be the same as those in VVLSVM and no further optimization was attempted.

\begin{figure}
\centering
\includegraphics[width=0.9\columnwidth]{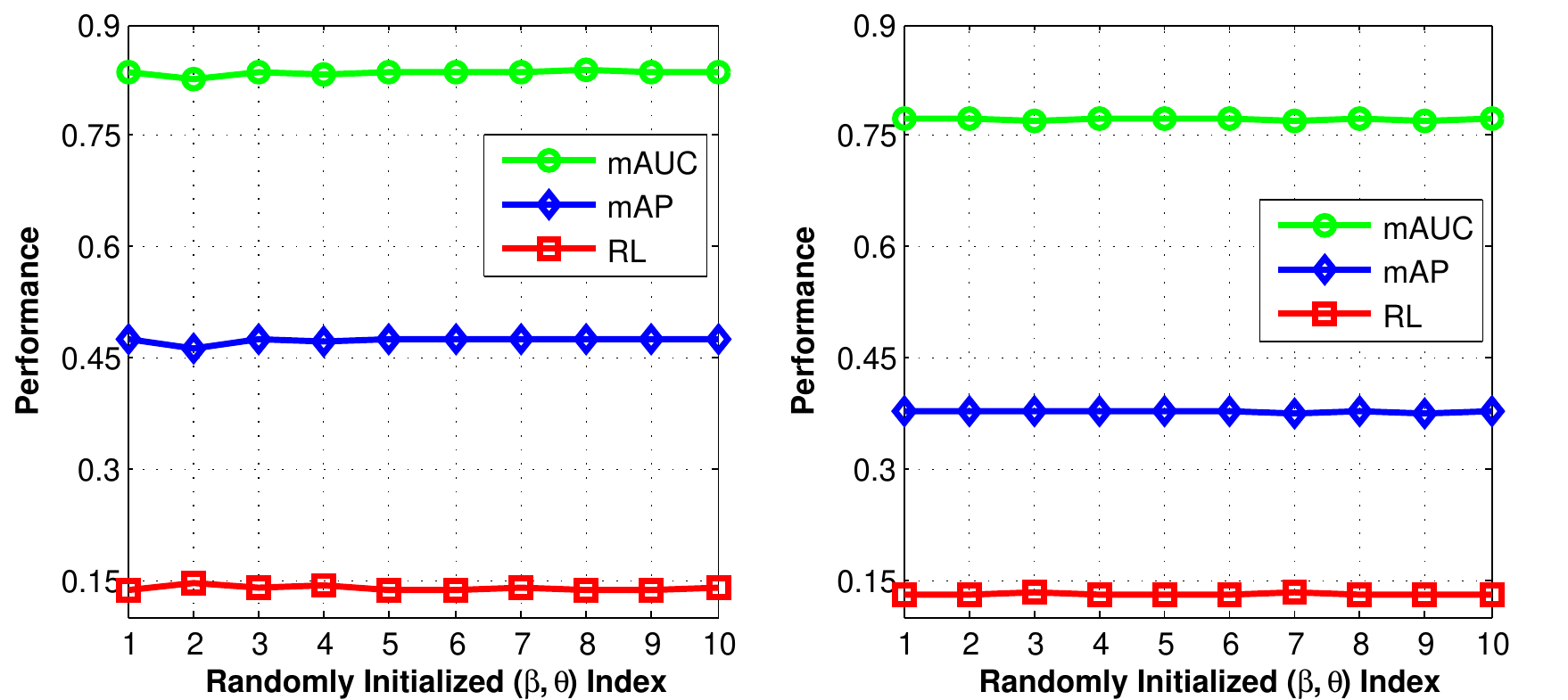}
\caption{Performance in mAP, mAUC, and RL vs. randomly initialized ($\beta$,$\theta$). The proposed model is insensitive to different initializations of ($\beta$,$\theta$). Left: PASCAL VOC' 07; Right: MIR Flickr.}
\label{fig:RandomInit}
\vspace{-5mm}
\end{figure}


The experimental results on the two datasets are shown in Fig. \ref{fig:Multiview_Enhancement}. We can see that learning the combination weights using our algorithm is always superior to simply using the uniform weights for different views. We also find that when the number of labeled samples increases, the improvement becomes small. This is because the multi-view learning actually helps to approximate the underlying data distribution. This approximation can be steadily improved with the increase of the number of labeled samples, and thus the significance of the multi-view learning to the approximation gradually decreases. Besides, we observe that $\beta$ has more influence on the final performance overall.

We show the behavior of the objective values by increasing the iteration number in Fig. \ref{fig:Objective_Value}. From the figure, we can see that only a few iterations (about five) are necessary to obtain a satisfactory solution. Thus the time complexity is only a little more than the VVLSVM algorithm and can justify the performance enhancement.

Finally, our algorithm is not sensitive to different initializations, as shown in Fig. \ref{fig:RandomInit}. In particular, we run our algorithm with 10 random choices of $\beta$ and $\theta$. We show the performance in terms of mAP, mAUC and RL on the two datasets in Fig. 6. It can be observed that the performance curves do not vary a lot with different initializations.

\subsection{Out-of-sample generalization}
\begin{figure}
\centering
\includegraphics[width=0.8\columnwidth]{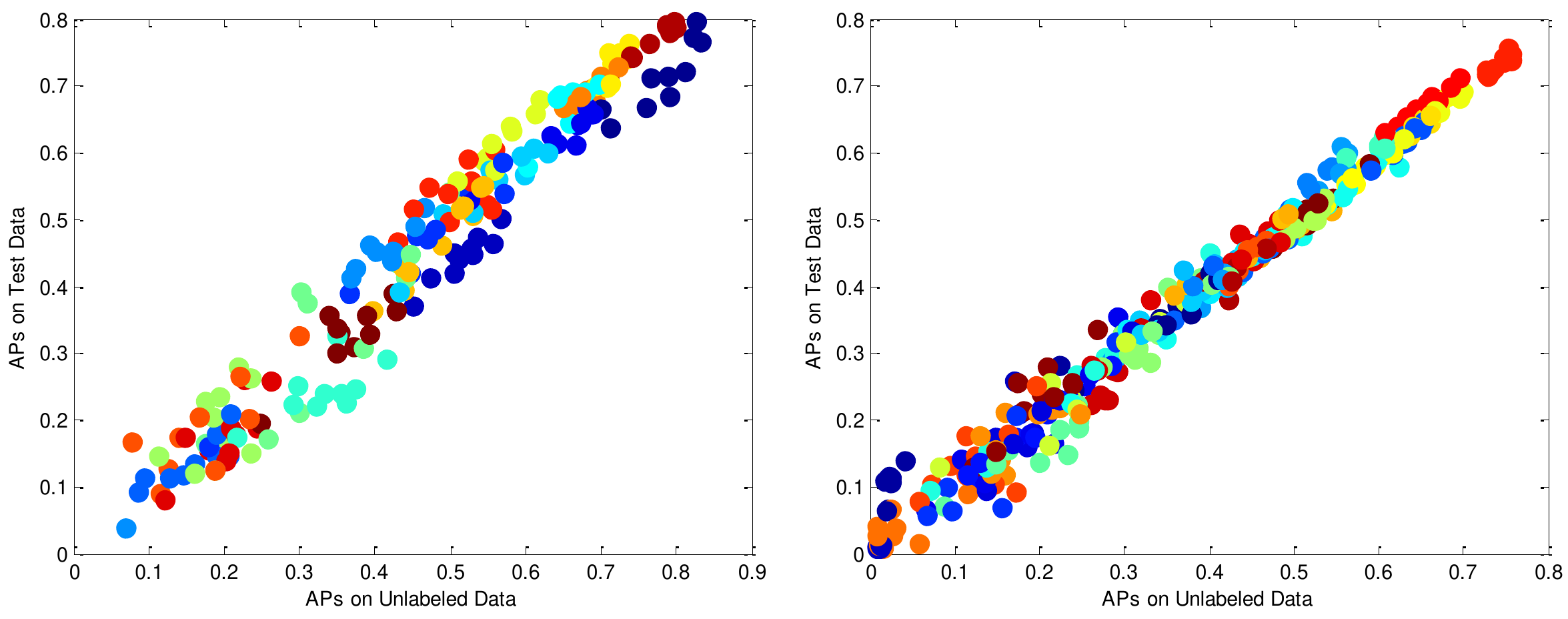}
\caption{Transductive and Inductive APs across outputs on VOC (Top) and MIR (Bottom) datasets.}
\label{fig:Out_Of_Sample_Extension}
\vspace{-5mm}
\end{figure}

The second set of experiments is to evaluate the out-of-sample extension quality of the MV$^3$MR framework and the SVM implementation is utilized. Fig. \ref{fig:Out_Of_Sample_Extension} compares the transductive performance to the inductive performance when using $l=200$ labeled samples. We show a scatter plot of the AP scores for each label on the two datasets by using 10-random choices of labeled data. We can see that our algorithm generalizes well from the unlabeled set to the unseen set. The MV$^3$MR framework inherits a strong natural out-of-sample generalization ability that many semi-supervised multi-label methods do not naturally have \cite{Minh-and-Sindhwani-ICML-2011}. Besides, most graph-based semi-supervised learning algorithms are transductive and additional induction schemes are necessary to handle new points \cite{Zhu-Madison-TR-2006}.

\subsection{Analysis of the combination coefficients in multi-view learning}
\begin{figure}
\centering
\includegraphics[width=1.0\columnwidth]{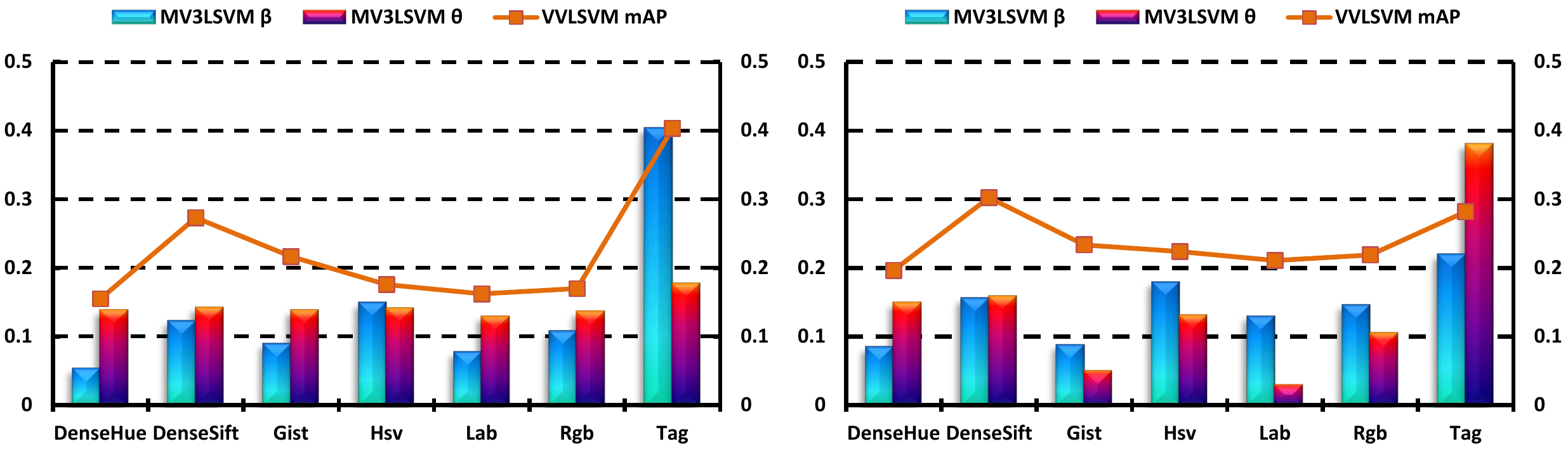}
\caption{The view combination weights $\beta$ and $\theta$ learned by MV$^3$MR, as well as the mAP of using VVLSVM for each view; Top: PASCAL VOC' 07; Bottom: MIR Flickr.}
\label{fig:Kernel_Combination}
\vspace{-5mm}
\end{figure}

In the following, we present empirical analyses of the multi-view learning procedure. In Fig. \ref{fig:Kernel_Combination}, we select $l=200$ and present the view combination coefficients $\beta$ and $\theta$ learned by MV$^3$LSVM, together with the mAP by using VVLSVM for each view. From the results, we find that the tendency of the kernel and graph weights are both consistent with the corresponding mAP in general, i.e., the views with a higher classification performance tend to be assigned larger weights, taking the DenseSIFT visual view (the 2nd view) and the tag (the last view) for example. However, a larger weight may sometimes be assigned to a less discriminative view; for example, the weight of Hsv (the 4th view) is larger than the weight of DenseSift (the 2nd view). This is mainly because the coefficient $\mathbf{a}$ is not optimal for every single view, in which only $G_v$ and $\mathcal{M}_v$ are utilized. The learned $\mathbf{a}$ minimizes the optimization problem (\ref{eq:MV3LSVM_Formulation}) by using the combined Gram matrix $G$ and integrated graph Laplacian $\mathcal{M}$, which means that the learned vector-valued function is smooth along the combined RKHS and the integrated manifold. In this way, the proposed algorithm effectively exploits the complementary property of different views.

\subsection{Comparisons with multi-label and multi-kernel learning algorithms}
Our last set of experiments is to compare MV$^3$LSVM with several competitive multi-label methods as well as a well-known MKL algorithm in predicting the unknown labels of the unlabeled data. The out-of-sample generalization ability of our method has been verified in our second set of experiments.

We specifically compare MV$^3$LSVM with the following methods on the challenging VOC and MIR datasets: \\
$\bullet$ \textbf{SVM\_CAT:} concatenating the features of each view and then running standard SVM. The parameter $C$ is tuned on the set $\{10^i|i=-1,0,\ldots,7,8\}$. The time complexity is $O(n l^{2.3})$ \cite{J-Platt-AKM-1999}. \\
$\bullet$ \textbf{SVM\_UNI:} combining different kernels by combining them with uniform weights and then running standard SVM. The parameter $C$ is tuned on the set $\{10^i|i=-1,0,\ldots,7,8\}$. The time complexity is $O(n l^{2.3})$ \cite{J-Platt-AKM-1999}. \\
$\bullet$ \textbf{MLCS \cite{Hsu-et-al-NIPS-2009}:} a multi-label compressed sensing algorithm that takes advantage of the sparsity of the labels. We choose the label compression ratio to be 1.0 since the number of the labels n is not very large. The mean of the multiple kernels from different views is pre-computed for experiments. Suppose the length the compressed label vector (for each sample) is $r \leq n$. Then the training cost is $O(n l^3)$ if we choose the regression algorithm to be the least squares \cite{M-Gori-WIRN-2011}, and the reconstruction complexity is $O(l(n^3 + r n^2))$ if the least angle regression (LARS) algorithm \cite{B-Efron-et-al-AoS-2004} is utilized. Considering that $r \leq n \leq l$ in this paper, the time complexity of MLCS is $O(n l^3)$. \\
$\bullet$ \textbf{KLS\_CCA \cite{Sun-et-al-PAMI-2011}:} a least-squares formulation of the kernelized canonical correlation analysis for multi-label classification. The ridge parameter is chose from the candidate set $\{0, 10^i|i=-3,-2,\ldots,2,3\}$. The mean of multiple Gram matrices is pre-computed to run the algorithm. According to the discussion presented in \cite{Sun-et-al-PAMI-2011}, the time complexity is $O(n^2 l+kn(3l+5d+2dl))$, where $d$ is the feature dimensionality and $k$ is the number of iterations. \\
$\bullet$ \textbf{SimpleMKL \cite{Rakotomamonjy-et-al-JMLR-2008}:} a popular SVM-based MKL algorithm that determines the combination of multiple kernels by a reduced gradient descent algorithm. The penalty factor $C$ is tuned on the set $\{10^i|i=-1,0,\ldots,7,8\}$. We apply SimpleMKL to multi-label classification by learning a binary classifier for each label. According to the Algorithm 1 presented in \cite{Rakotomamonjy-et-al-JMLR-2008}, there is an outer loop for updating the kernel weights, as well as an inner loop to determine the maximal admissible step size in the reduced gradient descent. Suppose the number of outer and inner iterations are $k_1$ and $k_2$ respectively, then the time complexity of SimpleMKL is approximately $O(n k_1 k_2 l^{2.3})$, where we have ignored the time cost of the SVM solver in the inner loop since it has warm start and can be very fast \cite{Rakotomamonjy-et-al-JMLR-2008}. \\
$\bullet$ \textbf{LpMKL \cite{Kloft-et-al-JMLR-2011}:} a recent proposed MKL algorithm, which extend MKL to $l_p$-norm with $p \geq 1$. The penalty factor $C$ is tuned on the set $\{10^i|i=-1,0,\ldots,7,8\}$ and we choose the norm $p$ from the set $\{1,8/7,4/3,2,4,8,16,\infty\}$. According to the Algorithm 1 presented in \cite{Kloft-et-al-JMLR-2011}, the time complexity is $O(n k l^{2.3})$ since the kernel combination coefficients can be computed analytically, where $k$ is the number of iterations.

\begin{table*}[!t]
\setlength\tabcolsep{3pt}
\renewcommand{\arraystretch}{1.3}
\caption{Performance evaluation on the two datasets}
\label{tab:Comparison_With_MLL_MKL}
\centering
\begin{tabular}{c|c|c|c||c|c|c|c}
\hline
\ & \multicolumn{3}{c||}{VOC \ \ mAP $\uparrow$ vs. \#\{labeled samples\}} & \multicolumn{3}{c|}{MIR \ \ mAP $\uparrow$ vs. \#\{labeled samples\}} & \ \\
\hline \hline
Methods & 100 & 200 & 500 & 100 & 200 & 500 & Ranks \\
\hline
SVM\_CAT & 0.241$\pm$0.011 (7) & 0.288$\pm$0.013 (7) & 0.371$\pm$0.007 (7)  & 0.281$\pm$0.009 (7) & 0.306$\pm$0.007 (7) & 0.352$\pm$0.008 (7) & 7 \\
SVM\_UNI & 0.347$\pm$0.018 (4.5) & 0.424$\pm$0.014 (5) & 0.529$\pm$0.006 (5)  & 0.302$\pm$0.011 (5) & 0.336$\pm$0.013 (6) & 0.400$\pm$0.009 (6) & 5.25 \\
MLCS \cite{Hsu-et-al-NIPS-2009} & 0.332$\pm$0.017 (6) & 0.412$\pm$0.016 (6) & 0.525$\pm$0.007 (6)  & 0.289$\pm$0.010 (6) & 0.342$\pm$0.011 (5) & 0.424$\pm$0.010 (5) & 5.67 \\
KLS\_CCA \cite{Sun-et-al-PAMI-2011} & 0.347$\pm$0.019 (4.5) & 0.432$\pm$0.014 (4) & 0.536$\pm$0.007 (4)  & 0.321$\pm$0.009 (3.5) & 0.369$\pm$0.017 (2) & 0.445$\pm$0.009 (2.5) & 3.42 \\
MV$^3$LSVM & \textbf{0.412$\pm$0.025 (1)} & \textbf{0.476$\pm$0.015 (1)} & \textbf{0.555$\pm$0.006 (1)}  & \textbf{0.332$\pm$0.013 (1)} & \textbf{0.376$\pm$0.017 (1)} & \textbf{0.449$\pm$0.008 (1)} & \textbf{1} \\
SimpleMKL \cite{Rakotomamonjy-et-al-JMLR-2008} & 0.381$\pm$0.024 (3) & 0.453$\pm$0.020 (3) & 0.538$\pm$0.011 (3)  & 0.321$\pm$0.014 (3.5) & 0.365$\pm$0.017 (4) & 0.444$\pm$0.011 (2.5) & 3.17 \\
LpMKL \cite{Kloft-et-al-JMLR-2011} & 0.391$\pm$0.024 (2) & 0.462$\pm$0.012 (2) & 0.540$\pm$0.006 (2) & 0.327$\pm$0.010 (2) & 0.367$\pm$0.014 (3) & 0.436$\pm$0.008 (4) & 2.5 \\
\hline \hline
\ & \multicolumn{3}{c||}{VOC \ \ mAUC $\uparrow$ vs. \#\{labeled samples\}} & \multicolumn{3}{c|}{MIR \ \ mAUC $\uparrow$ vs. \#\{labeled samples\}} & \ \\
\hline \hline
Methods & 100 & 200 & 500 & 100 & 200 & 500 & Ranks \\
\hline
SVM\_CAT & 0.744$\pm$0.013 (7) & 0.785$\pm$0.006 (7) & 0.832$\pm$0.003 (7)  & 0.722$\pm$0.008 (4) & 0.745$\pm$0.004 (6) & 0.783$\pm$0.004 (6) & 6.17 \\
SVM\_UNI & 0.783$\pm$0.008 (3) & 0.824$\pm$0.009 (2.5) & 0.870$\pm$0.003 (2.5)  & 0.718$\pm$0.011 (5) & 0.742$\pm$0.011 (7) & 0.782$\pm$0.006 (7) & 4.5 \\
MLCS \cite{Hsu-et-al-NIPS-2009} & 0.773$\pm$0.010 (5) & 0.819$\pm$0.010 (6) & 0.869$\pm$0.004 (4)  & 0.701$\pm$0.012 (7) & 0.749$\pm$0.010 (5) & \textbf{0.805$\pm$0.005 (1.5)} & 4.42 \\
KLS\_CCA \cite{Sun-et-al-PAMI-2011} & 0.781$\pm$0.009 (4) & 0.824$\pm$0.008 (2.5) & 0.866$\pm$0.003 (5)  & 0.737$\pm$0.009 (2) & \textbf{0.769$\pm$0.010 (1.5)} & \textbf{0.805$\pm$0.005 (1.5)} & 2.75 \\
MV$^3$LSVM & \textbf{0.801$\pm$0.011 (1)} & \textbf{0.835$\pm$0.011 (1)} & \textbf{0.875$\pm$0.004 (1)}  & \textbf{0.741$\pm$0.014 (1)} & \textbf{0.769$\pm$0.012 (1.5)} & 0.802$\pm$0.005 (3.5) & \textbf{1.5} \\
SimpleMKL \cite{Rakotomamonjy-et-al-JMLR-2008} & 0.769$\pm$0.017 (6) & 0.822$\pm$0.013 (4.5) & 0.870$\pm$0.006 (2.5)  & 0.717$\pm$0.013 (6) & 0.753$\pm$0.010 (4) & 0.802$\pm$0.005 (3.5) & 4.42 \\
LpMKL \cite{Kloft-et-al-JMLR-2011} & 0.786$\pm$0.008 (2) & 0.822$\pm$0.008 (4.5) & 0.862$\pm$0.005 (6) & 0.732$\pm$0.010 (3) & 0.756$\pm$0.010 (3) & 0.795$\pm$0.007 (5) & 3.92 \\
\hline \hline
\ & \multicolumn{3}{c||}{VOC \ \ RL $\downarrow$ vs. \#\{labeled samples\}} & \multicolumn{3}{c|}{MIR \ \ RL $\downarrow$ vs. \#\{labeled samples\}} & \ \\
\hline \hline
Methods & 100 & 200 & 500 & 100 & 200 & 500 & Ranks \\
\hline
SVM\_CAT & 0.220$\pm$0.008 (7) & 0.183$\pm$0.006 (7) & 0.142$\pm$0.003 (7)  & 0.165$\pm$0.005 (3) & 0.146$\pm$0.004 (5) & 0.126$\pm$0.002 (5) & 5.67 \\
SVM\_UNI & 0.178$\pm$0.008 (2) & 0.143$\pm$0.006 (3) & \textbf{0.106$\pm$0.003 (1.5)}  & 0.549$\pm$0.040 (7) & 0.437$\pm$0.022 (7) & 0.177$\pm$0.011 (7) & 4.58 \\
MLCS \cite{Hsu-et-al-NIPS-2009} & 0.195$\pm$0.007 (5) & 0.155$\pm$0.007 (5) & 0.112$\pm$0.004 (4)  & 0.173$\pm$0.006 (5) & 0.145$\pm$0.005 (4) & 0.115$\pm$0.003 (2.5) & 4.25 \\
KLS\_CCA \cite{Sun-et-al-PAMI-2011} & 0.183$\pm$0.008 (3) & 0.149$\pm$0.006 (4) & 0.122$\pm$0.005 (5)  & 0.168$\pm$0.013 (4) & 0.143$\pm$0.005 (3) & 0.121$\pm$0.004 (4) & 3.83 \\
MV$^3$LSVM & \textbf{0.170$\pm$0.007 (1)} & \textbf{0.140$\pm$0.007 (1)} & 0.108$\pm$0.003 (3)  & \textbf{0.150$\pm$0.006 (1)} & \textbf{0.130$\pm$0.007 (1)} & \textbf{0.111$\pm$0.003 (1)} & \textbf{1.33} \\
SimpleMKL \cite{Rakotomamonjy-et-al-JMLR-2008} & 0.214$\pm$0.017 (6) & 0.142$\pm$0.009 (2) & \textbf{0.106$\pm$0.004 (1.5)}  & 0.155$\pm$0.005 (2) & 0.136$\pm$0.006 (2) & 0.115$\pm$0.003 (2.5) & 2.67 \\
LpMKL \cite{Kloft-et-al-JMLR-2011} & 0.186$\pm$0.011 (4) & 0.164$\pm$0.010 (6) & 0.137$\pm$0.007 (6) & 0.199$\pm$0.014 (6) & 0.181$\pm$0.007 (6) & 0.141$\pm$0.004 (6) & 5.67 \\
\hline
\end{tabular}
\begin{tabular}{m{1.9\columnwidth}}
($\uparrow$ indicates ``the larger the better''; $\downarrow$ indicates ``the smaller the better''. Mean and std. are reported. The best result is highlighted in boldface.)
\end{tabular}
\end{table*}

The performance of the compared methods on the VOC dataset and MIR dataset are reported in Table \ref{tab:Comparison_With_MLL_MKL}. The values in the last column of Table \ref{tab:Comparison_With_MLL_MKL} are average ranks. From the results, we firstly observe that the performance keeps improving with the increasing number of the labeled samples. Second, the performance of the simpleMKL algorithm, which learns the kernel weights for SVM, can be inferior to the multi-label algorithms with the mean kernel in many cases. MV3LSVM is superior to multi-view (SimpleMKL and LpMKL) and multi-label algorithms in general and consistently outperforms other methods in terms of mAP. The average rank of our algorithm is smaller than all the other methods in terms of all the three criteria. According to the Friedman test \cite{Demsar-JMLR-2006}, the statistics $F_F$ of mAP, mAUC and RL are $56.05$, $3.03$, and $5.69$ respectively. All of them are larger than the critical value $F(6,30)=2.42$, so we reject the null-hypothesis (the compared algorithms perform equally well). In particular, by comparing with SimpleMKL, we obtain a significant $8.1\%$ mAP improvement on VOC when using 100 labeled samples. The level of improvement drops when more labeled samples are available, for the same reason described in our first set of experiments.

\section{Conclusion and Discussion}
\label{sec:Conclusion}
Most of the existing works on multi-label image classification use only single feature representation, and the multiple feature methods usually assume that a single label is assigned to an image. However, an image is usually associated with multiple labels and different kinds of features are necessary to describe the image properly. Therefore, we have developed a multi-view vector-valued manifold regularization (MV$^3$MR) for multi-label image classification in which images are naturally characterized by multiple views. MV$^3$MR combines different kinds of features in the learning process of the vector-valued function for multi-label classification. We also derived an SVM formulation of MV$^3$MR, which results in MV$^3$LSVM. The new algorithm effectively exploits the label correlations and learns the view weights to integrate the consistency and complementary properties of different views. We evaluate the proposed algorithm in terms of three popular criteria, i.e. mAP, mAUC and RL. Intensive experiments on two challenge datasets PASCAL VOC'07 and MIR Flickr show that the support vector machine based implementation under MV$^3$MR outperforms the traditional multi-label algorithms as well as a well-known multiple kernel learning method. Furthermore, our method provides a strategy for learning from multiple views in multi-label classification and can be extended to other multi-label algorithms.

\appendices
\section{PROOF OF LEMMA \ref{lm:M_Graph_Laplacian}}
\begin{IEEEproof}
The matrix $\mathcal{M} = \sum_v \theta_v \mathcal{M}_v = \sum_v \theta_v (\mathcal{L}_v \otimes I_n) = \mathcal{L} \otimes I_n$, where $\mathcal{L} = \sum_v \theta_v \mathcal{L}_v$ is defined as a convex combination of the scalar-valued graph Laplacians constructed from different views. $\mathcal{L} \in S_N^+$ since each $\mathcal{L}_v \in S_N^+$, and thus we have $\mathcal{M} \in S_{Nn}^+$ according to the positive definite property on the Kronecker product. Here, $\mathcal{L}=\sum_v \theta_v (\mathcal{D}_v-\mathcal{W}_v)$ can be computed by using the following adjacency graph
\begin{numcases}
{\mathcal{W}_{ij}=}{}
\begin{split}
\notag
\sum_v \theta_v \mathcal{W}_{vij} \ & \mathrm{if} \ x_i \in N(x_j) \ \mathrm{or} \ x_j \in N(x_i), \\
\notag
0 \ \ \ \ \ \ \ & \mathrm{otherwise},
\end{split}
\end{numcases}
where $N(x)$ denotes a set that contains the $k$-nearest neighbors of $x$ and $\mathcal{W}_{vij}$ is the similarity between the $i$th and $j$th point from the $v$th view. Thus $\mathcal{L}$ is a graph Laplacian and $\mathcal{M}$ is the corresponding vector-valued graph Laplacian.
\end{IEEEproof}

\section{PROOF OF THE REPRESENTER THEOREM}
\begin{IEEEproof}
It has been presented in Section \ref{subsec:Rationality} that there is an RKHS $\mathcal{H}_K$ associated with the vector-valued kernel $K$. The probability distribution is assumed to be supported on a manifold $M$ in the manifold regularization framework. We now denote $S=\{\sum_i K(x_i,\cdot)a_i | x_i \in M,a_i \in \mathcal{Y}\}$ as a linear space spanned by the kernels centered at the points on $M$. Any function $f \in \mathcal{H}_K$ can be decomposed as $f = f_\parallel + f_\perp$, with $f_\parallel \in S$ and $f_\perp \in S^\perp$. It has been proved in Lemma \ref{lm:M_Graph_Laplacian} that $\mathcal{M}$ is a graph Laplacian. Thus we can use $\mathcal{M}$ to induce an intrinsic norm $\|\cdot\|_I$, which satisfies $\|f\|_I=\|g\|_I$ for any $f,g \in \mathcal{H}_K$, $(f-g)|_M \equiv 0$. According to the reproducing property, it concludes that $f_\perp$ vanishes on $M$ \cite{Belkin-et-al-JMLR-2006}. This means that for any $x \in M$, we have $f(x)=f_\parallel (x)$ and then $\|f\|_I=\| f_\parallel \|_I$. Besides $\|f\|_K^2=\| f_\parallel \|_K^2 + \| f_\perp \|_K^2 \geq \| f_\parallel \|_K^2$, and thus we conclude that the minimizer of the problem (\ref{eq:MV3MR_Formulation}) must lie in $S$ for fixed $\beta$ and $\theta$. Furthermore, because $M$ is approximated by the Laplacian of the graph constructed by the labeled and unlabeled samples, we have $S=\{\sum_{i=1}^{l+u} K(x_i,\cdot) a_i | a_i \in \mathcal{Y}\}$. This completes the proof.
\end{IEEEproof}



\ifCLASSOPTIONcaptionsoff
  \newpage
\fi



\scriptsize
\bibliographystyle{IEEEtran}
\bibliography{./TNNLS-2012-P-0852}
\end{document}